\pdfoutput=1
\documentclass[11pt]{article}

\usepackage{EMNLP2022}

\usepackage{times}
\usepackage{latexsym}
\usepackage{hyperref}

\usepackage[T1]{fontenc}
\usepackage[utf8]{inputenc}

\usepackage{microtype}

\usepackage{inconsolata}

\usepackage{times}
\usepackage{supertabular}
\usepackage{longtable}
\usepackage{latexsym}
\usepackage{amsmath}
\usepackage{mathabx}
\usepackage{booktabs} % To thicken table lines
\usepackage{multicol}
\usepackage{slashbox}
\usepackage[T1]{fontenc}
\newcommand{\wx}[1]{\textcolor{magenta}{\bf\small [#1 --WX]}}

\usepackage[utf8]{inputenc}

\usepackage{soul}
\usepackage{tikz}
\usepackage{longtable}
\usepackage[export]{adjustbox}
\usepackage{subcaption}
\usepackage{makecell}

\usepackage{cleveref}
\usepackage{xcolor, color}
\usepackage{graphicx}
\usepackage{multirow}
\usepackage{booktabs}
\usepackage{comment}
\usepackage{float}
\usepackage{enumitem}
\usepackage{pbox}
\usepackage{anyfontsize} % for changing font size of table 7
\usepackage{microtype}

\definecolor{lightblue}{RGB}{219,226,238}
\definecolor{darkred}{RGB}{105,17,10}
\definecolor{lightyellow}{RGB}{251,242,214}
\definecolor{darkyellow}{RGB}{82,58,34}
\definecolor{lightgrey}{RGB}{230,230,230}
\definecolor{darkgrey}{RGB}{57,57,57}
\definecolor{lightgreen}{RGB}{152,251,152}
\definecolor{darkgreen}{RGB}{34,139,34}
\definecolor{lightpurple}{RGB}{200, 140, 200}
\definecolor{darkpurple}{RGB}{160, 0, 190}

\newcommand{\stanceosaurus}{Stanceosaurus}

\usepackage{tabularx,booktabs}
\newcolumntype{Y}{>{\centering\arraybackslash}X}

\usepackage{array}
\newcolumntype{P}[1]{>{\raggedright\arraybackslash}p{#1}}
\newcolumntype{C}[1]{>{\centering\arraybackslash}m{#1}}

\usepackage{devanagari}
\usepackage{scalerel,xparse}

\newcolumntype{M}[1]{>{\centering\arraybackslash}m{#1}}

\NewDocumentCommand\emojisauropod{}{
    \includegraphics[scale=0.09]{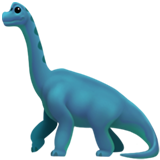}
}

\title{\emojisauropod {\sc Stanceosaurus}: Classifying Stance Towards \\ Multicultural Misinformation}

\author{Jonathan Zheng, Ashutosh Baheti, Tarek Naous, Wei Xu, Alan Ritter \\
  School of Interactive Computing \\
  Georgia Institute of Technology \\
  \small{
 \texttt{\{jonathanqzheng, abaheti3,  tareknaous\}@gatech.edu; \{wei.xu, alan.ritter\}@cc.gatech.edu}}
\\}

\begin{document}
\maketitle
\begin{abstract}

% Existing stance datasets only focus on misinformation claims originating from one region, example Western Countries or Arab Countries, thus ignoring the multi-cultural aspect of misinformation spread.
We present \textit{\stanceosaurus}, a new corpus of 28,033 tweets in English, Hindi, and Arabic annotated with stance towards 251 misinformation claims.  As far as we are aware, it is the largest corpus annotated with stance towards misinformation claims.  The claims in {\stanceosaurus} originate from 15 fact-checking sources that cover diverse geographical regions and cultures.
Unlike existing stance datasets, we introduce a more fine-grained 5-class labeling strategy with additional subcategories to distinguish implicit stance.
Pretrained transformer-based stance classifiers that are fine-tuned on our corpus show good generalization on unseen claims and regional claims from countries outside the training data.
% with 61.0 F1 and 57.1 F1 respectively
Cross-lingual experiments demonstrate {\stanceosaurus}' capability of training multilingual models, achieving 53.1 F1 on Hindi and 50.4 F1 on Arabic without any target-language fine-tuning. Finally, we show how a domain adaptation method can be used to improve performance on {\stanceosaurus} using additional RumourEval-2019 data. We make {\stanceosaurus} publicly available to the research community and hope it will encourage further work on misinformation identification across languages and cultures.\footnote{Our code and data are available at \url{https://tinyurl.com/stanceosaurus}}

\end{abstract}

\section{Introduction}

\begin{table*}[t]
\setlength{\tabcolsep}{4pt}%Tighter
\centering
% \small
\small
\resizebox{0.98\textwidth}{!}{%
\begin{tabular}{lcl}
\toprule
\textbf{Dataset} & \textbf{Target}  & \textbf{Number/Range of Topics}  \\
\midrule
SemEval-2016 \cite{mohammad-etal-2016-semeval} & Subject  & 6 political topics (e.g., \textit{atheism}, \textit{feminist movement}) \\

SRQ \cite{villacox2020stance} & Subject  & 4 political topics \& events (e.g., \textit{general terms}, \textit{student marches}) \\

Catalonia \cite{zotova-etal-2020-multilingual} & Subject & 1 topic (i.e., \textit{Catalonia independence}) \\

COVID \cite{glandt-etal-2021-stance} & Subject  & 4 topic related to Covid-19 (e.g., \textit{stay at
home orders})\\

Multi-target \cite{sobhani-etal-2017-dataset} & Entity  & 3 pairs of candidates in 2016 US election \\

WTWT \cite{conforti-etal-2020-will} & Event & 5 merger and acquisition events\\

RumourEval \cite{gorrell-etal-2019-semeval} & Tweet & 8 news events + rumors about natural disasters \\

Rumor-has-it \cite{qazvinian-etal-2011-rumor} & Claim & 5 rumors (e.g., \textit{ Sarah Palin getting divorced?}) \\
%\midrule
%Perspectrum \citep{chen-etal-2019-seeing} & & & \\

CovidLies \cite{hossain-etal-2020-covidlies} & Claim & 86 pieces of COVID-19 misinformation \\
\midrule
%AR: Not twitter...
%AraStance \cite{alhindi-etal-2021-arastance} & Claim  & 910 claims from fact checking websites in Aarabic speaking countries.  \\
%\midrule
Stanceosaurus (this work) & Claim & \textbf{251} claims over a diverse set of global and regional topics \\
\bottomrule
\end{tabular}
}
\vspace{-.1in}
\caption{Summary of Twitter stance classification datasets. {\stanceosaurus} covers more claims from a broader range of topics and geographical regions than prior Twitter stance datasets.}
\vspace{-.4cm}
\label{tab:survey} 
\end{table*}
%Online misinformation has become a serious problem, prompting social media websites such as Facebook and Twitter to spend billions of dollars on content moderation.\footnote{\url{https://www.cnbc.com/2021/02/27/content-moderation-on-social-media.html}} 

The prevalence of misinformation on online social media has become an increasingly severe societal problem. A key language technology, which has the potential to help content moderators identify rapidly-spreading misinformation, is the automatic identification of both affective and epistemic stance \citep{jaffe2009stance,zuczkowski2017epistemic} towards false claims.  Progress on the problem of stance identification has largely been driven by the availability of annotated corpora, such as RumourEval \citep{derczynski-etal-2017-semeval, gorrell-etal-2019-semeval}.  However, existing corpora mostly focus on misinformation spreading within western countries.

\begin{figure}[bt]
    \centering
    \includegraphics[width=0.98\linewidth]{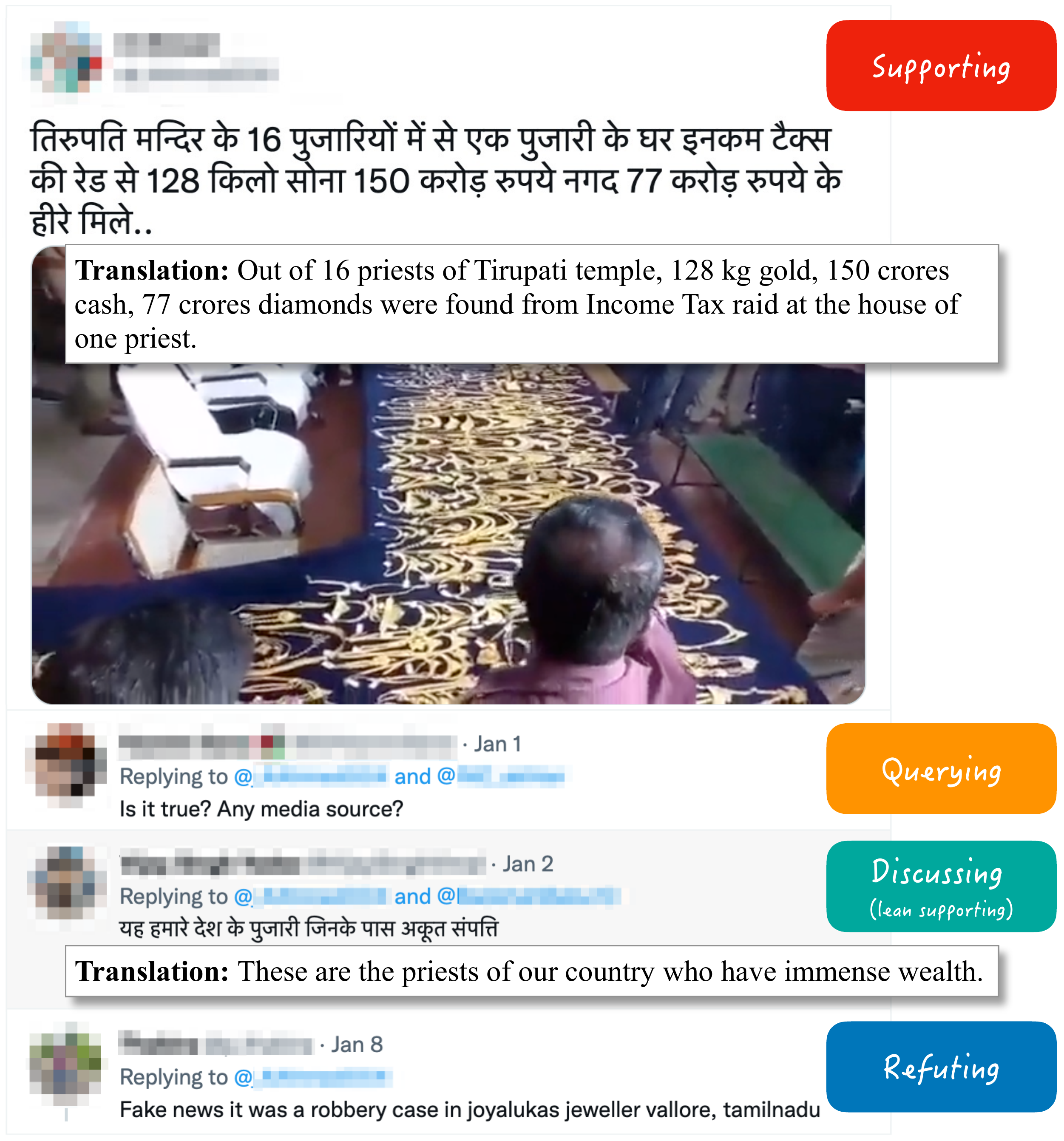}
    \caption{Example Hindi and English tweets in Stanceosaurus with stance towards the claim {\em ``Raid at Tirupati temple priest's house, 128 kg gold found''}.}
    \label{fig:data_collection}
\end{figure}

%each has specific limitations.   
%For instance, RumourEval annotates stance of replies in a conversation thread towards an earlier utterance, limiting its utility for identification of misinformation outside a specific reply chain. 

%The first limitation in prior work is that, existing datasets are limited in terms of the breadth of topics covered.  For example, RumourEval 2019 subtask A (henceforth, RumourEval) consists of English tweets and reply threads, annotated with stance towards the source tweet, and covers eight topics related to events in western countries.  Another limitation is that... {\bf AR: TODO} \cite{jaffe2009stance}

In this paper, we present {\em Stanceosaurus}, a diverse and high-quality corpus that builds on the best design choices made in previous misinformation corpora, including RumourEval-2019 and CovidLies.
%are annotated with epistemic stance towards fact-checked claims. 
Stanceosaurus covers more diverse topics, geographic regions, and cultures than prior work. It includes 28,033 tweets in English, Hindi, and Arabic that are manually annotated for stance (see Figure \ref{fig:data_collection}) towards 251 misinformation claims, collected from 15 independent fact-checking websites that cover India, Singapore, Australia, New Zealand, Canada, the United States, Europe, and the Arab World (the Levantine, Gulf, Northwest African regions, and Egypt). To the best of our knowledge, {\stanceosaurus} is the largest and most diverse annotated stance dataset to date.

Through extensive experiments, we demonstrate that {\stanceosaurus} can support the fine-grained classification of explicit and implicit stances, as well as zero-shot cross-lingual stance identification. In addition, we introduce and experiment with class-balanced focal loss \cite{cui2019class} to alleviate the class imbalance issue, which is a well-known challenge in automatic stance detection \cite{zubiaga-etal-2016-stance,baly2018integrating}. Similar to other corpora that are labeled with stance towards messages or claims, {\stanceosaurus} reflects the natural distribution of stance observed {\em in the wild}, with comparatively few examples labeled as Supporting or Refuting (see label distributions in Table \ref{tab:dataset_stats}). We show that fine-tuning BERTweet$_{large}$ with class-balanced focal loss \cite{cui2019class} can achieve 66.8 F1 for 3-way stance classification and 61.0 F1 for the finer-grained 5-way stances for English. With zero-shot transfer learning, we achieve 50.4 and 53.1 F1 for Hindi and Arabic, respectively, in a 5-way classification. Lastly, we show it is possible to train a single model to achieve better performance on Stanceosaurus' test set via additional fine-tuning on RumourEval \cite{gorrell-etal-2019-semeval}, using a variation of EasyAdapt \cite{daume2007frustratingly,bai-etal-2021-pre} designed for pre-trained Transformers, even though these two corpora have significant differences.

\section{Related Work}
\setlength{\tabcolsep}{3pt}%Tighter
\begin{table*}[!ht]
\centering
\small
\resizebox{0.98\textwidth}{!}{
\begin{tabular}{llcccccccc}
\toprule
\textbf{Source} & \textbf{Country \& Regions} & \textbf{Lang} & \textbf{\#Claims} & \textbf{\#Tweets} & \textbf{\underline{\smash{Irr.}}} & \textbf{\underline{\smash{Sup.}}} & \textbf{\underline{\smash{Ref.}}} & \textbf{\underline{\smash{Dis.}}} & \textbf{\underline{\smash{Que.}}} \\ \midrule
Snopes & USA (80\%), INT'L (16.7\%), Other (3.3\%) & en & 30 & 3197  & 1051 & 428 & 229 & 1447 & 42 \\ 
Poynter & Europe (5\%), INT'L (90\%), Other (5\%) & en & 20 & 2197 & 949 & 274 & 97 & 844 & 33 \\
FullFact & UK (30\%), INT'L (55\%), Other (15\%) & en & 20 & 2379 & 806 & 300 & 179 & 1057 & 37 \\
AFP Fact Check CAN & Canada (55\%), INT'L (30\%), Other (15\%)  & en & 20 & 2078 & 746 & 252 & 130 & 910 & 40 \\
AAP Fact Check & Australia (10\%), INT'L (65\%), Other (25\%) & en & 20 & 2302 & 739 & 374 & 136 & 1019 & 34 \\
AFP Fact Check NZ & New Zealand (15\%), INT'L (75\%), Other (10\%) & en & 20 & 2227 & 879 & 194 & 81 & 1044 & 29 \\
Blackdotresearch & Singapore (30\%), INT'L (55\%), Other (15\%) & en & 20 & 2307 & 842 & 248 & 113 & 1076 & 28 \\
Factly & India (45\%), INT'L (55\%) & en & 20 & 1979 & 889 & 190 & 117 & 734 & 49 \\
Politifact & USA (20\%), INT'L (35\%), Other (45\%) & en & 20 & 2041 & 984 & 289 & 8 & 753 & 7 \\
% Alt News & India (90\%), INT'L (10\%)  & hi & 20 & 737 & 125 & 269 & 68 & 266 & 9 \\
% Alt News & India (90\%), INT'L (10\%)  & hi & 20 & 1437 & 381 & 416 & 159 & 464 & 17 \\
Alt News & India (90.4\%), INT'L (4.8\%), Other (4.8\%)  & hi & 21 & 1730 & 550 & 489 & 172 & 500 & 19 \\
Aajtak & India (67\%), Other (33\%)  & hi & 9 & 806 & 456 & 110 & 40 & 193 & 7 \\
Hindi Newschecker & India (56\%), Other (44\%)  & hi & 9 & 781 & 195 & 313 & 46 & 219 & 8 \\
MISBAR & Arab World (58.3\%), INT'L (8.3\%), Other (33.4\%) & ar & 12 & 2283 & 454 & 514 & 203 & 1031 & 81 \\
Fatabyyano & Arab World (28.5\%), INT'L (57.1\%), Other (14.4\%) & ar & 7 & 986 & 234 & 163 & 49 & 522 & 18 \\
Maharat Fact-o-meter & INT'L (100\%) & ar & 3 & 740 & 224 & 132 & 55 & 316 & 13 \\
\midrule
% Total & & & 170 & 18666 & 7759 & 2516 & 1275 & 6844 & 268 \\
Total & Regional (57.4\%), INT'L (42.6\%) & & 251 & 28033 & 9998 & 4270 & 1655 & 11665 & 445 \\
\bottomrule
\end{tabular}
}
\caption{Fact-checking sources included in our {\stanceosaurus} corpus. The most common regions are listed. \textbf{Stance} -- breakdown of tweets into 5 main categories in relation to each claim: \underline{\smash{Irr}}elevant, \underline{\smash{Sup}}porting, \underline{\smash{Ref}}uting, \underline{\smash{Dis}}cussing, and \underline{\smash{Que}}rying. \textbf{Country \& Regions} -- home country of the source and the distribution of claims regarding home country, regional, and international matters. Other refers to claims in countries other than the primary countries covered by the source (e.g., Snopes claims about India).}
% \wx{Add Arabic sources, enlarge Hindi; include some true claims for English/Arabic.}
\label{tab:corpus_stats}
\end{table*}

\textbf{Stance Classification Datasets.} Given the importance of studying misinformation spreading on Twitter and the open access to its data, there are many stance classification datasets consisting of annotated tweets. However, existing datasets are largely restricted to a limited range and a number of topics --- see Table \ref{tab:survey} for a summary.\footnote{See also the excellent survey by  \citeauthor{DBLP:journals/corr/abs-2103-00242} \shortcite{DBLP:journals/corr/abs-2103-00242}. Given space limitations, we highlight only the most relevant work.} Note that many of these datasets are considering stance toward an entity or topic (e.g., \textit{Bitcoin}), whereas we focus on more specific full-sentence claims (e.g., \textit{Bitcoin is legal in Malaysia}), which provides flexibility to cover more diverse topics in our work. 

The closet prior efforts to ours are RumourEval-2019 \cite{gorrell-etal-2019-semeval} and CovidLies \cite{hossain-etal-2020-covidlies}. RumourEval-2019 \cite{gorrell-etal-2019-semeval} contains annotations on whether a reply tweet in a conversation thread is supporting, denying, querying, or commenting on the rumour mentioned in the source tweet. However, RumourEval covers only eight major news events (e.g., \textit{Charlie
Hebdo shooting}) plus additional rumors about natural disasters. The CovidLies dataset \cite{hossain-etal-2020-covidlies} annotates a 3-way stance (Agree, Disagree, Neutral) towards 86 pieces of COVID-19-specific misinformation, using BERTScore \cite{zhang2019bertscore} to find potentially relevant tweets.  As the authors of CovidLies \cite{hossain-etal-2020-covidlies} have noted, relying on BERTScore (i.e., a semantic similarity measurement) biases the data collection towards more supporting and less refuting tweets. %We combined and improved upon the best practices from both works, namely considering reply chains as in RumourEval-2019 and full-sentence claims as in CovidLies. We further improved the data annotation schema and expanded the coverage to more regional and diverse topics.  

Besides Twitter, stance classification has also been studied for other types of data. For example, the Perspectrum dataset \cite{chen-etal-2019-seeing} was constructed using debate forum data. Emergent \cite{ferreira-vlachos-2016-emergent} and AraStance \cite{alhindi-etal-2021-arastance} consist of English and Arabic news articles annotated with stance, respectively. %In contrast, {\stanceosaurus} covers a broader set of geographic regions and supports cross-lingual stance identification experiments. 

\paragraph{Fact Checking Datasets.} Related to but different from stance classification, fact-checking (aka rumour verification) as an NLP task primarily focuses on the assessment of claims being true or false. There exist several fact-checking datasets, such as FEVER \cite{thorne-etal-2018-fever} and MultiFC \cite{augenstein-etal-2019-multifc} for English, and X-Fact \cite{gupta-srikumar-2021-x} for 25 non-English languages. These datasets consist of claims extracted from Wikipedia or fact-checking sites, which are labeled for veracity. 

%, without needing the efforts of manually annotating individual tweets or text snippets as in our and other works that create the stance classification datasets. 

%The feasibility for an academic research group to create multilingual datasets 

\paragraph{Automatic Stance Classification.} Many prior efforts have developed methods for automatic stance classification, which have progressed from feature-based approaches \cite{qazvinian-etal-2011-rumor, lukasik-etal-2015-classifying, ferreira-vlachos-2016-emergent, zeng2016unconfirmed, aker-etal-2017-simple, riedel2017simple, DBLP:conf/www/ZhangYL18, ghanem-etal-2018-stance, 10.1145/3295823, li-etal-2019-eventai} to neural approaches \cite{kochkina-etal-2017-turing, chen-etal-2017-ikm, 10.1145/3132847.3133116, DBLP:conf/www/BhattSSNRM18, hanselowski-etal-2018-retrospective, 8576019, 9178321}, then to fine-tuning of pre-trained models \cite{10.1007/978-3-030-28577-7_4, fajcik-etal-2019-fit, matero-etal-2021-melt-message}. Researchers \cite{zubiaga-etal-2016-stance} have noted the class imbalance issue in stance classification and subsequently chose Macro F1 as the main evaluation metric. To deal with imbalanced data, previous works have used methods such as per-label weights \cite{garcia-lozano-etal-2017-mama, ghanem-etal-2019-upv}, oversampling underrepresented examples \cite{singh-etal-2017-iitp}, retrieving additional examples from external datasets \cite{yang-etal-2019-blcu}, or adjusting prediction thresholds over class label probabilities \cite{li-scarton-2020-revisiting}. With the availability of many small-scale stance datasets, other works attempted weakly supervised \cite{Kumar2020, yang2022weakly}, semi-supervised methods \cite{8528884}, or multi-task models \cite{kochkina-etal-2018-one, DBLP:conf/www/MaGW18, li-etal-2019-rumor-detection, wei-etal-2019-modeling, kumar-carley-2019-tree, fang-etal-2019-neural, cheng2020vroc, yu-etal-2020-coupled, 10.1145/3430984.3431007}. A few efforts have also looked at transferring knowledge from larger datasets to smaller datasets \cite{Xu2019AdversarialDA, hardalov-etal-2021-cross, schiller2021stance} and languages with less data \cite{mohtarami-etal-2019-contrastive, zotova-etal-2020-multilingual, hardalov2021few}. {\stanceosaurus} (this work) is one of the largest and most diverse stance classification datasets to date, enabling the study of cross-lingual transfer for stance classification towards misinformation claims.

%In contrast, we use the class-balanced focal loss \cite{cui2019class} to address the data imbalance problem which naturally assigns more weight to difficult training examples. 

%It worth mentioning that, due to the limited amount of annotated data that exist, many previous works resorted to leveraging additional data from other related tasks, such as sentiment and emotion detection, to improve the performance. 

%\section{{\stanceosaurus} Data Collection}
\section{The {\stanceosaurus} Corpus}
\label{sec:data_coll}

Our corpus consists of social media posts manually annotated for stance toward claims from multiple fact-checking websites across the world. We carefully designed the data collection and annotation scheme to ensure better quality and coverage, improving upon prior work.

\subsection{Collecting Fact-checked Claims}
% Diversity in previous stance datasets
% multi-cultural and borad coverage of regional and international news
%Prior stance classification datasets have limited their data sources to popular breaking news and celebrity rumours, restricting diversity.

To ensure multicultural representation, we obtain fact-checked claims from both Western and non-Western sources (Table \ref{tab:corpus_stats}). We choose nine well-known fact-checking websites in English, three in Hindi, and three in Arabic.\footnote{Fact-checking sources are selected from \href{https://en.wikipedia.org/wiki/List_of_fact-checking_websites}{Wikipedia}, \href{https://ifcncodeofprinciples.poynter.org/signatories}{Poynter's International Fact-Checking Network}, as well as those in X-Fact \cite{gupta-srikumar-2021-x}.} We randomly select claims from each source posted between 5/17/2012 and 02/28/2022 that have sparked discussion on Twitter. In total, we have 251 claims in our corpus, of which 144 are considered regional based on manual inspection (see column \textbf{Country \& Regions} in Table \ref{tab:corpus_stats}). For example, the claims {\em ``Finland is promoting a 4 day work week''} and {\em ``Burning Ghee will produce Oxygen''}\footnote{Ghee is a type of clarified butter, commonly used in cuisines from the Indian subcontinent.} are both considered regional, one explicitly and one implicitly; whereas the claim {\em ``Bees use acoustic levitation to fly''} is considered international. The claims in Stanceosaurus range from news, health, and science to politics (e.g., {\em ``Sonu Sood promises to support Hamas/Palestine''}), conspiracy theories, history, and urban myths (e.g., {\em ``The pyramids of Giza were built by slaves''}). We present all 251 claims in Appendix \ref{sec:claims}.

\subsection{Retrieving Conversations around Claims}
\label{sec:retrieving}

For better coverage of diverse topics, we invested substantial effort in creating customized queries with varied keywords and time ranges for each claim to retrieve tweets. We also trace the entire reply chain in both directions, so \stanceosaurus{} includes relevant tweets that may not contain the keywords.
%are also included in our dataset. 

%\footnote{Some other prior works on Twitter-based stance datasets used hashtags, such as ``\#covid'' or ``\#IranNuclearDeal'' to retrieve tweets \cite{mohammad-etal-2016-semeval}.}

%Following prior works \cite{qazvinian-etal-2011-rumor,gorrell-etal-2019-semeval,derczynski-etal-2017-semeval}

\paragraph{Curated Search Queries.} We retrieve tweets by keyword search, which we believe is the most effective approach given the constraints of Twitter's APIs. To ensure the coverage and quality of our dataset, we manually curated and iteratively refined search queries for each claim, utilizing advanced search operators to restrict the relevant time period and language. We expand search queries with synonyms (e.g., {\em ``jab''} for {\em ``vaccine''}) and lexical variations whenever possible; the latter is particularly helpful for including different Arabic dialects. See Appendix \ref{appendix:claims_and_queries} for example queries. 
%One seemingly small but important design choice worth mentioning is that we deliberately relax our search queries to obtain some lexically similar but irrelevant tweets, which allows training of more robust stance classifiers that can directly operate over raw Twitter data. 
We collect tweets from different time periods for different claims (e.g., a two-week range for timely events and a max range from 7/3/2008 to 5/9/2022 for historic myths). %To include different dialects in Arabic, we designed multiple search queries per claim with different synonyms and lexical variations. 

% \textcolor{red}{1/5/2022} \wx{update for Arabic}. 
%For example, we use a time period from 2020 to present for COVID related claims, an unrestricted time range for historic myths, and a couple of weeks after the event for claims about political slander, disasters or school shootings. 
%While keyword search is a costly step that involves manual efforts as in this work, it is also one of the most feasible and reliable approaches given the constraints of Twitter APIs. 

%The time range used for the search queries vary from claims to claims. 

%In contrast, most previous stance datasets only consider comments directly related to a topic or claim. Using our search queries, we sample up to 50 tweets per claim. 

%We use the Twitter search API to retrieve messages related to the 190 fact-checked claims. Search queries were manually curated by combining relevant keywords and iteratively refined to improve coverage.
% Existing stance classification datasets only allowed comments relevant to a topic or claim.

% 50 tweets using search queries + 50 parent tweets + 50 reply tweets = at most 150 tweets per claim
% A complete list of the claims and their corresponding queries is presented in Table \ref{tab:query_claim_english} and Table \ref{tab:query_claim_hindi}\ab{For myself - Convert this to a single table/figure with few example claims and queries} (see Appendix \ref{appendix:claims_and_queries}).

%\label{sec:context}
\paragraph{Context from URLs and Reply Chains.}
Individual tweets retrieved by search do not capture the contextual aspects of stance, which can be very important as misinformation often spreads in multi-turn conversations on social media.
Therefore, we also collect the parent tweets (i.e., the tweet that a search retrieved tweet is replying to) and the entire reply chains if available. 
% From these, we sample 50 parent tweets and 50 reply chain tweets. 
Additional details are presented in Appendix \ref{app:conversation_retrieval}.

\subsection{Annotating Stance Towards Claims}
\label{sec:stance_categories}

We employ a fine-grained annotation scheme that supports 5-way and 3-way stance classification.

% After collecting the raw data as described in \S \ref{sec:retrieving}, we randomly sampled 50 tweets per claim for annotation.
%After we retrieved the data points from our list of selected topics, we sample 50 tweets per claim to populate the dataset, and we manually annotate the sample for its stance.  We identify 50 data points per claim to be a healthy balance between a large enough dataset and a dataset that requires too much time to manually annotate.
%\subsection{Stance Categories}

\paragraph{5-way Stance Categories.}

We define stance detection as a five-way classification task, including irrelevant tweets in addition to the four stance classes used in prior works \cite{schiller2021stance, gorrell2018rumoureval}, as follows:
\begin{itemize}[leftmargin=18pt]
\itemsep-.5em
\item \textbf{Irrelevant} -- unrelated to the claim;
\item \textbf{Supporting} -- explicitly affirms the claim is true or provides verifying evidence;
\item \textbf{Refuting} -- explicitly asserts the claim is false or presents evidence to disprove the claim;
\item \textbf{Discussing} -- provide neutral information on the context or veracity of the claim;
\item \textbf{Querying} -- questions the veracity of the claim.
\end{itemize}

\noindent See Figure \ref{fig:data_collection} and Appendix \ref{app:example_replychain} for examples of different stances, shown with the reply chain details.

%\wx{what does the following sentence mean?} When the parent tweet is available in the context, we propagate the stance of previous tweets downwards along with the context to make accurate judgments. Table \ref{tab:tweet_examples} provides examples of \textbf{context} and \textbf{contextless} data, with context-dependent tweets demarcated.

% \wx{We could be somewhat more brief about these subcategories in the main paper, and move the detailed definition of each subcategory into the Appendix? keep the high-level idea in main. The idea is that the main paper is very compact and condensed, so that you keep reader's interest of reading it. It could be a bit tedious for the reader to read and distinguish all these many subcategories, so we shall pick the most interesting points and give high level ideas. }

\paragraph{Subcategories and 3-way Stance Classification.} 
Although some tweets may be neutral towards a claim, they can still show an indirect bias. For example, the tweet {\em ``Fauci: No Concern About Number of People Testing Positive After COVID-19 Vaccine.''} in response to the claim {\em ``The COVID-19 Vaccine has magnets or will make your body magnetic''} discusses the vaccine rollout, while it can be viewed as implicitly supporting the claim regarding the lack of vaccine safety. We thus further annotate the Discussing tweets for their leanings as three subcategories: Discussing$_{support}$ (44.6\%),  Discussing$_{refute}$ (25.7\%), and Discussing$_{other}$ (29.7\%). This not only enables fine-grained classification but also makes our {\stanceosaurus} corpus flexible enough to support the 3-way (Supporting, Refuting, Other)\footnote{By merging (1) Discussing$_{support}$ with Supporting; (2) Discussing$_{refute}$ with Refuting;  (3) Discussing$_{other}$, Irrelevant, and Querying together into Other.} setup used in other prior work.

\paragraph{Data Annotation.}
We hired four native speakers for English, two for Hindi, and two for Arabic to annotate the tweets with stance. English annotators are all from the U.S., and non-English annotators grew up in the respective countries or regions of the claims being collected. All of the annotators have a college-level education. We designed detailed guidelines (see Appendix \ref{subsec:tricky_anno}) and held training sessions to assist our annotators. For each claim, the annotators are reasonably familiar with the topic because they are asked to read and learn about the subject matter before annotating. Cohen's Kappa ($\kappa$) between the annotators is summarized in Table \ref{tab:kappa}, showing substantial agreement \citep{artstein2008inter} for all languages. 

\begin{table}[h]
\centering
\small
\begin{tabular}{rccc}
\toprule
  \textbf{\#Tweets} & \textbf{Lang} & \textbf{5-class $\kappa$} & \textbf{3-class $\kappa$}\\ 
\midrule
 20,707 & en & 0.624 & 0.670 \\
 3,317 & hi & 0.673 & 0.742 \\
 4,009 & ar & 0.773 & 0.729 \\

\bottomrule
\end{tabular}
\caption{Inter-annotator agreement calculated based on 5-class and 3-class stances.}
\label{tab:kappa}
\end{table}

Disagreements often occur over challenging cases. For example, 
%{\em ``Evergreen ship stuck in the Suez Canal - interesting call sign URL''} 
{\em ``Evergreen ship stuck in the Suez Canal - interesting call sign''} 
is supporting the conspiracy theory {\em ``Hillary Clinton is trafficking children aboard the Evergreen Ship''}, with the connection being that the call sign of the ship is {\em ``H3RC''}, which coincidentally overlaps with Hillary's initials. The disagreements were resolved by a third adjudicator for Hindi, and through discussions between the annotators for English and Arabic. Interestingly, the Hindi subset of {\stanceosaurus} exhibits some forms of code-switching in 28.2\% of instances, including some replies written in English, while 6.3\% of the Arabic data exhibited code-switching. A subset of 200 tweets randomly sampled from the Arabic data was further labeled for language variations, which contains 62.5\% Modern Standard Arabic (MSA), 35.5\% dialects, 0.5\% Arabizi, and 1.5\% in the form of emojis or mentions.

\subsection{Comparison to RumourEval}
% RumourEval is most comparable to our work
% Main distinctions - No explicit claims, irrelevant tweets, leanings
% Difference from the original set (added URL context)
Although our annotation design is comparable to RumourEval \cite{gorrell-etal-2019-semeval}, in that both annotate the stance of Twitter threads towards rumorous claims, there are a few important differences: (1) RumourEval limits their rumorous claims primarily to 8 major news events plus additional natural disaster events, whereas we use a much larger and more diverse sample of claims originating from multicultural news outlets. (2) RumourEval, unlike our dataset, does not explicitly provide the claims. Rather, the first tweet of the thread is used to represent both the claim and the stance in RumourEval. (3) We label discussing subcategories that capture indirect bias towards a claim (see \S \ref{sec:stance_categories}). (4) RumourEval excludes irrelevant tweets, limiting its generalizability. For a direct comparison, we present the corpus statistics of Stanceosaurus and RumourEval-2019\footnote{As RumourEval distributes only message IDs, we reconstructed the dataset by retrieving all the available posts from Twitter and Reddit, with a loss of a small portion of data that has been deleted on the social media platform (120 out of 1876 instances in the test set; 12 and 7 instances in the train/dev).} in Table \ref{tab:dataset_stats}, and further test classification models on both datasets in \S \ref{sec:domain_adaptation_rumoureval}.

\section{Automatic Stance Detection}

% \wx{may need to update} 
We design multiple automatic stance identification experiments to test the generalization capabilities of models trained on {\stanceosaurus}. First, we establish the baseline performance of predicting stance towards unseen claim using fine-tuned Transformer models in \S \ref{sec:unseen_claims} and experiment with the class-balanced focal loss for addressing the imbalanced class distribution. We present zero-short cross-lingual experiments in \S \ref{subsec:hindi_classification}, where multilingual models are trained on English tweets and evaluated on the Hindi and Arabic tweets. Furthermore, we demonstrate that a simple domain adaptation method can help improve performance on {\stanceosaurus} using additional RumourEval data in \S \ref{sec:domain_adaptation_rumoureval}. Finally, we show that models trained only on International claims subset can extrapolate well to regional claims from individual countries in \S \ref{sec:unseen_countries}.

\begin{table}[t]
\centering
\small
\resizebox{0.48\textwidth}{!}{
\begin{tabular}{l|rrr||rrr}
\toprule
 \multirow{2}{*}{\textbf{Stance}} & \multicolumn{3}{c||}{\textbf{\stanceosaurus}} &  \multicolumn{3}{c}{\textbf{RumourEval}} \\
%& \textbf{Train} & \textbf{Test} & \textbf{Validation} & \textbf{Train} & \textbf{Test} & \textbf{Validation} & \textbf{Train} & \textbf{Test} & \textbf{Validation} 

&\textbf{\#train} & \textbf{\#test} & \multicolumn{1}{c||}{\textbf{\#dev}} & \textbf{\#train} & \textbf{\#test} & \textbf{\#dev}
\\ \midrule
Irrelevant & 4928 & 1674 & 1283 \hspace*{1mm} & --- & --- &  --- \\ 
Supporting & 1462 & 592 & 495 \hspace*{1mm} & 925 & 157 & 102 \\
Refuting & 598 & 270 & 222 \hspace*{1mm} & 378 & 101 & 82 \\
Discussing & 4941 & 2160 & 1783 \hspace*{1mm} & 3507 & 1405 & 1174 \\
$~+$Other & 949 & 532 & 366 \hspace*{1mm} & --- & --- & --- \\
$~+$Supporting & 2440 & 1082 & 780 \hspace*{1mm} & --- & --- & --- \\
$~+$Refuting & 1552 & 546 & 637 \hspace*{1mm} & --- & --- & --- \\
Querying & 201 & 54 & 44 \hspace*{1mm} & 395 & 93 & 120 \\
\midrule
\rule{0pt}{2ex} Total & 12130 & 4750 & 3827 \hspace*{1mm} & 5205 & 1756 & 1478 \\
\bottomrule
\end{tabular}
} %(end resizebox)
\caption{(Left) Number of tweets in the English subset of {\stanceosaurus}. The train/dev/test sets consist of 112/44/34 separate claims, respectively. (Right) Statistics of RumourEval-2019 \cite{gorrell-etal-2019-semeval} after we reconstruct the data from message IDs.}
\label{tab:dataset_stats}
\end{table}

\subsection{Baseline Models}
%  To demonstrate that Stanceosaurus supports training and evaluating stance identification models, and , 

We experiment with fine-tuning methods using BERT  \cite{devlin-etal-2019-bert} and BERTweet \cite{nguyen-etal-2020-bertweet}. The latter is a RoBERTa-based \cite{DBLP:journals/corr/abs-1907-11692} model pre-trained on Twitter data.\footnote{The $base$ size of the BERTweet model is trained on 850M English tweets streamed from 01/2012 to 08/2019. The $large$ size is trained with additional 23M tweets that are related to COVID-19.} Stance identification is modeled as sentence-pair classification, using special tokens to format the input as ``[CLS] claim [SEP] text'', where ``text'' is a tweet concatenated with its context (parent tweet and any extracted HTML titles -- see \S \ref{sec:retrieving}). We found that incorporating context generally helps stance classification for reply tweets (see ablation study in Appendix \ref{subsec:context_ablation}). We use standard cross-entropy loss in all baselines. %\wx{We moved the experiments about context (root/reply tweets) to appendix, so we need to add a brief mention about it somewhere, and smooth out the section.}  

\subsection{Class-balanced Focal Loss (CB$_{\text{foc}}$)}

\label{sec:cb_focal_loss}
%One common challenge in stance classification is the class imbalance problem.
The imbalanced class problem has been identified as a major challenge in automatic stance classification \cite{li-scarton-2020-revisiting}, since fewer messages exhibit Supporting or Refuting stances in the wild (see Table \ref{tab:dataset_stats}). To alleviate this issue, prior work has used weighted cross-entropy loss \cite{fajcik-etal-2019-fit}. We experiment with weighted cross-entropy loss and Class-Balanced Focal loss \cite{cui2019class,baheti-etal-2021-just}, which has shown promising results in computer vision research recently, as an alternative.

%this loss on Stanceosaurus and RumourEval

% To address the challenge of imbalanced label distributions in Stanceosaurus and RumourEval we 

%which also prevails in {\stanceosaurus} due to its low proportion of naturally occurring supporting and refuting tweets (14\% and 7\%, respectively). Therefore, in addition to the standard cross-entropy loss function, we also experiment with class-balanced focal loss \cite{cui2019class} to address the class imbalance problem.
% \cite{li-scarton-2020-revisiting}. To elucidate, In the 5-way stance classification, supporting and refuting stance naturally occur only in a small portion of data (14\% and 7\%, respectively).
% , and thus difficult to be identified by automatic classifiers. %Most of the tweets are either irrelevant (even though containing the keywords) or discussions (even though leaning towards supporting or refuting implicitly).

We use $\hat{s} = (z_0, z_1, z_2, z_3, z_4) $ to represent the unnormalized scores assigned by the model for five stance classes $C = $ \{Irrelevant, Discussing, Supporting, Refuting, Querying\}. The class-balanced focal loss is then defined as:
\[
\text{CB\textsubscript{foc}}(\hat{s}, y) = - \underbrace{\frac{1 - \beta}{1 - \beta^{n_y}}}_\text{reweighting} \underbrace{\sum_{m \in C} (1 - p_m)^{\gamma} \log(p_m)}_\text{focal loss}.
\]
\noindent
$y$ is the gold stance label, $n_y$ is the number of instances with the label $y$, and $p_m = sigmoid(z'_m)$, where:
\[z'_m = \begin{cases}
      z_m & m = y \\
      -z_m & \text{otherwise}
    \end{cases}
\]
\noindent
Focal loss employs the expression $(1-p_m)^\gamma$ to reduce the relative loss for well classified examples \cite{lin2017focal}. The reweighting term lowers the impact of class imbalance on the loss. In our experiments, hyperparameters $\beta$ and $\gamma$ are tuned between [0.1, 1) and [0.1, 1.1], respectively, based on the performance on the dev set. 
%[0.1, 0.9999] and [0.1,1.1], respectively. 

\subsection{Implementation Details} 
We replace usernames and URLs with special tokens, truncate or pad the input to a sequence length of 256 as in BERT and BERTweet.\footnote{We use the maximum sequence length of 128 tokens for BERTweet$_{base}$.}
%\wx{we replace the website URLs, but still consider website titles -- did I get it correctly? in later experiment about adding context, we may need to separate the website titles as an additional row in the table to justify this decision} 
%For BERTweet, normalizing the usernames and website urls ended up hurting the model performance, so the text was not normalized \wx{a bit weired that that normalizing username and urls hurt performance for BERTweet, as its authors seem to recommend doing so}. 
All models were trained for 10 epochs and optimized with the Adam optimizer. Learning rates were selected among $\{1e^{-5}$, $3e^{-5}$, $5e^{-5}$, $7e^{-5}$, $9e^{-5}\}$. The train batch size was set to 8. For all test set evaluations, we select the best checkpoint that achieves the highest Macro F1 on the development set.

% \begin{table*}
% \centering
% \small
% \begin{tabular}{l|ccc|ccc|ccc|ccc}
% \toprule
%  \rule{0pt}{2.6ex} \multirow{2}{*}{\textbf{Model}} & \multicolumn{3}{c|}{\textbf{Context}} & \multicolumn{3}{c|}{\textbf{Standalone}} & \multicolumn{3}{c}{\textbf{Trained Standalone}} & \multicolumn{3}{|c}{\textbf{Decontextualized}} \\
% & \textbf{Pres} & \textbf{Recall} & \textbf{F1} & \textbf{Pres} & \textbf{Recall} & \textbf{F1} & \textbf{Pres} & \textbf{Recall} & \textbf{F1} & \textbf{Pres} & \textbf{Recall} & \textbf{F1} \\ \toprule
% BERT & 51.4 & \textbf{52.5} & \textbf{51.5} & \textbf{62.8} & 58.3 & 60.0 & 58.2 & 60.1 & 58.4 & 54.7 & 54.9 & 54.1 \\
    % BERTweet & \textbf{52.1} & 51.1 & 49.9 & 60.1 & \textbf{61.7} & \textbf{59.9} & \textbf{59.3} & \textbf{60.9} & \textbf{58.9} & \textbf{55.4} & 54.2 & 54.2 \\
% BERT-Dual & 49.8 & 50.5 & 49.9 & 58.9 & 59.2 & 59.0 & 58.8 & 59.8 & 59.1 & 54.7 & \textbf{55.5} & \textbf{54.5} \\

% \bottomrule

% \end{tabular}
% \caption{\label{Dataset Table}
% Table of the context (Unseen Claims setup; )
% Context (Trained on entire dataset, tested on only the context examples of the test set)
% Standalone (Trained on entire dataset, tested on only the standalone examples of the test set)
% Trained standalone (Trained on standalone examples on training set, tested on only the standalone examples of the test set)
% Decontextualized (Trained on all data without context, tested on the entire test set without context as well); 'context' -- appended the tweet before.
% }
% \end{table*}

\begin{table}[b!]
\centering
\small
\renewcommand{\arraystretch}{0.85}% Tighter
\resizebox{0.48\textwidth}{!}{
\begin{tabular}{l|ccc}
\toprule
  \multirow{2}{*}{\textbf{Model}} & \multicolumn{3}{c}{\textbf{Stanceosaurus \scalebox{0.75}{(unseen claims)}}} \\
& \hspace*{1mm} \textbf{Precision} & \hspace*{2mm} \textbf{Recall} & \textbf{F1} \\ \midrule
BERT$_{base}$+ CE & 51.1{\scriptsize $\pm$1.1} & 50.5{\scriptsize $\pm$2.0} & 50.4{\scriptsize $\pm$1.6} \\
+ weighted CE & 50.5{\scriptsize $\pm$1.9} & 52.7{\scriptsize $\pm$1.1} & 51.3{\scriptsize $\pm$1.3} \\
+ CB$_{\text{foc}}$ & 50.6{\scriptsize $\pm$1.3} & 55.7{\scriptsize $\pm$2.1} & 52.5{\scriptsize $\pm$1.0} \\
\midrule
BERT$_{large}$+ CE & 54.3{\scriptsize $\pm$0.8} & 53.0{\scriptsize $\pm$0.6} & 53.6{\scriptsize $\pm$0.6} \\
+ weighted CE & 53.8{\scriptsize $\pm$1.3} & 53.8{\scriptsize $\pm$1.2} & 53.6{\scriptsize $\pm$1.0} \\
+ CB$_{\text{foc}}$ & 53.9{\scriptsize $\pm$1.2} & 53.7{\scriptsize $\pm$1.1} & 53.6{\scriptsize $\pm$0.5} \\
\midrule
BERTweet$_{base}$+ CE & 53.1{\scriptsize $\pm$1.2} & 52.2{\scriptsize $\pm$1.6} & 52.3{\scriptsize $\pm$1.0} \\
+ weighted CE & 51.8{\scriptsize $\pm$1.0} & 55.2{\scriptsize $\pm$1.4} & 53.1{\scriptsize $\pm$0.7} \\
+ CB$_{\text{foc}}$ & 51.3{\scriptsize $\pm$0.6} & 56.8{\scriptsize $\pm$0.6} & 53.5{\scriptsize $\pm$0.3} \\ 
\midrule
BERTweet$_{large}$+ CE & 60.6{\scriptsize $\pm$2.0} & 60.2{\scriptsize $\pm$1.0} & 60.2{\scriptsize $\pm$1.1} \\
+ weighted CE & \textbf{60.8}{\scriptsize $\pm$1.6} & 60.2{\scriptsize $\pm$1.0} & 60.2{\scriptsize $\pm$0.5} \\
+ CB$_{\text{foc}}$ & 59.8{\scriptsize $\pm$1.3} & \textbf{62.8}{\scriptsize $\pm$1.5} & \textbf{61.0}{\scriptsize $\pm$0.8} \\
\bottomrule
\end{tabular}
} %(end resizebox)
\caption{5-way stance classification results for unseen claims in {\stanceosaurus} (mean$\pm$standard deviation across runs of five random seeds). Class-balanced focal loss (CB$_{\text{foc}}$) outperforms standard and weighted cross-entropy loss (CE, weighted CE).}
\label{tab:main_results}
\end{table}

\begin{table*}[t]
\centering
\renewcommand{\arraystretch}{0.95}% Tighter
\setlength{\tabcolsep}{1.5pt}%Tighter
\scriptsize
% \begin{tabular}{l|l|r|M{1.2cm}M{1.2cm}M{1.2cm}|M{1.2cm}M{1.2cm}M{1.2cm}}
\resizebox{\textwidth}{!}{
%\begin{tabular}{l|l|r|M{1.1cm}M{1.1cm}M{1.1cm}|M{1.1cm}M{1.1cm}M{1.1cm}|M{1.1cm}M{1.1cm}M{1.1cm}}
\begin{tabular}{l|l|r|M{1cm}M{1cm}M{1cm}|M{1cm}M{1cm}M{1cm}|M{1cm}M{1cm}M{1cm}}
\toprule
\multicolumn{2}{c|}{\multirow{2}{*}{\textbf{Stance Class}}} & \multirow{2}{*}{\textbf{\#test}} & \multicolumn{3}{c|}{\textbf{Cross-Entropy Loss}} & \multicolumn{3}{c|}{\textbf{Weighted Cross-Entropy Loss}} & \multicolumn{3}{c}{\textbf{Class-balanced Focal Loss}} \\
\multicolumn{2}{c|}{} & \multicolumn{1}{c|}{} & \textbf{Precision} & \textbf{Recall} & \textbf{F1} & \textbf{Precision} & \textbf{Recall} & \textbf{F1} & \textbf{Precision} & \textbf{Recall} & \textbf{F1} \\ \midrule
\multirow{7}{*}{5-Class} & Supporting & 592 & \textbf{59.9}{\tiny $\pm$2.1} & 61.5{\tiny $\pm$2.2} & \textbf{60.6}{\tiny $\pm$1.0} & 57.2{\tiny $\pm$0.9} & 60.6{\tiny $\pm$2.3} & 58.8{\tiny $\pm$1.6} & 57.6{\tiny $\pm$1.0} & \textbf{63.8}{\tiny $\pm$1.3} & 60.5{\tiny $\pm$1.0} \\
& Refuting & 270 & 60.6{\tiny $\pm$6.9} & 57.4{\tiny $\pm$1.9} & 58.7{\tiny $\pm$2.3} & 60.6{\tiny $\pm$3.0} & \textbf{61.6}{\tiny $\pm$4.6} & 60.9{\tiny $\pm$1.1} & \textbf{60.9}{\tiny $\pm$2.2} & \textbf{61.6}{\tiny $\pm$3.2} & \textbf{61.1}{\tiny $\pm$1.0} \\
& Discussing & 2160 & 66.4{\tiny $\pm$0.7} & 63.5{\tiny $\pm$2.7} & \textbf{64.9}{\tiny $\pm$1.3} & 65.5{\tiny $\pm$1.1} & \textbf{64.1}{\tiny $\pm$2.9} & 64.7{\tiny $\pm$1.0} & \textbf{67.0}{\tiny $\pm$1.1} & 60.0{\tiny $\pm$1.8} & 63.2{\tiny $\pm$0.8} \\ 
& Querying & 54 & 43.7{\tiny $\pm$7.2} & 42.6{\tiny $\pm$5.9} & 42.5{\tiny $\pm$3.7} & \textbf{47.5}{\tiny $\pm$6.8} & 41.1{\tiny $\pm$4.3} & 43.6{\tiny $\pm$2.7} & 42.3{\tiny $\pm$5.7} & \textbf{51.5}{\tiny $\pm$6.6} & \textbf{45.8}{\tiny $\pm$2.9} \\
& Irrelevant & 1674 & 72.4{\tiny $\pm$2.4} & 76.0{\tiny $\pm$2.3} & \textbf{74.1}{\tiny $\pm$0.9} & \textbf{73.2}{\tiny $\pm$2.3} & 73.4{\tiny $\pm$5.0} & 73.2{\tiny $\pm$1.5} & 71.1{\tiny $\pm$0.9} & \textbf{77.4}{\tiny $\pm$2.8} & \textbf{74.1}{\tiny $\pm$0.9} \\ \cmidrule{2-12}
& All & 3912 & 60.6{\tiny $\pm$2.0} & 60.2{\tiny $\pm$1.0} & 60.2{\tiny $\pm$1.1} & \textbf{60.8}{\tiny $\pm$1.6} & 60.2{\tiny $\pm$1.0} & 60.2{\tiny $\pm$0.5} & 59.8{\tiny $\pm$1.3} & \textbf{62.8}{\tiny $\pm$1.5} & \textbf{61.0}{\tiny $\pm$0.8} \\
\midrule
\multirow{5}{*}{3-Class} & Supporting & 1674 & 66.9{\tiny $\pm$1.6} & 68.1{\tiny $\pm$1.3} & 67.5{\tiny $\pm$1.3} & 68.9{\tiny $\pm$3.1} & \textbf{68.2}{\tiny $\pm$4.6} & \textbf{68.3}{\tiny $\pm$1.3} & \textbf{70.1}{\tiny $\pm$1.5} & 65.0{\tiny $\pm$2.9} & 67.4{\tiny $\pm$1.0} \\
& Refuting & 816 & 55.2{\tiny $\pm$1.8} & 51.9{\tiny $\pm$4.5} & 53.3{\tiny $\pm$1.6} & \textbf{56.0}{\tiny $\pm$2.5} & 52.2{\tiny $\pm$2.4} & 53.9{\tiny $\pm$1.5} & 54.5{\tiny $\pm$3.0} & \textbf{58.5}{\tiny $\pm$5.1} & \textbf{56.2}{\tiny $\pm$0.8} \\
& Other & 2260 & 75.9{\tiny $\pm$1.3} & 76.4{\tiny $\pm$1.2} & 76.1{\tiny $\pm$0.5} & 75.9{\tiny $\pm$1.9} & \textbf{77.9}{\tiny $\pm$2.9} & 76.8{\tiny $\pm$0.6} & \textbf{76.2}{\tiny $\pm$1.6} & \textbf{77.9}{\tiny $\pm$1.6} & \textbf{77.0}{\tiny $\pm$0.2} \\ 
\cmidrule{2-12}
 & All & 3912 & 66.0{\tiny $\pm$0.4} & 65.5{\tiny $\pm$1.1} & 65.6{\tiny $\pm$0.7} & 66.9{\tiny $\pm$0.9} & 66.1{\tiny $\pm$1.0} & 66.4{\tiny $\pm$0.9} & \textbf{66.9}{\tiny $\pm$0.5} & \textbf{67.1}{\tiny $\pm$0.9} & \textbf{66.8}{\tiny $\pm$0.4} \\
\bottomrule

\end{tabular}
}
\caption{
Per-label comparison of BERTweet$_{large}$, when fine-tuned with cross-entropy, weighted cross-entropy loss, and class-balanced focal loss, both for 3-class and 5-class stance detection on our corpus. Weighted cross-entropy and class-balanced focal loss improves F1 score overall, and in particular for the least frequent stance of Refuting.
%For all but the two most common labels (Irrelevant and Discussing), training with class-balanced focal loss improves the F1 score.
}
\label{tab:3or5-class}
\end{table*}

\section{Experiments and Results}

We report average results over five random seeds primarily by Macro F1, which has been used as the standard metric in stance classification since the arguably more important stances (i.e., Refuting and Supporting) only consist of a small portion of data. %Precision@k, an alternative to Macro F1, can also be a useful classifier metric, as it could provide extra information on how the classifier would perform in downstream applications, such as helping content moderators flag or delete misinformation.

\subsection{Stance Detection for Unseen Claims}
\label{sec:unseen_claims}
For this experiment, we split the English data based on claims into train, dev, and test set (see the left side of Table \ref{tab:dataset_stats}). We evaluate all models on the 5-way stance classification of tweets towards claims that are unseen during training. As shown in Table \ref{tab:main_results}, the best model is BERTweet$_{large}$, which achieves 60.2 F1 when trained with standard and weighted cross-entropy loss and 61.0 F1 with class-balanced focal loss. We see some alleviation of the data imbalance issue in the per-label analysis in Table \ref{tab:3or5-class}, which shows improved F1 using class-balanced focal loss for the two least frequent labels, Refuting and Querying.

%We train models with different loss functions on this 5-way stance classification setup and report Macro F1 scores. 

%treats all classes equally by averaging per class F1, which 

%() 

% 

%7\% and 14\% of data, respectively

%\footnote{To ensure higher quality data, about half of the tweets in dev and test set were annotated by a second annotator. The disagreements in these new annotations were resolved with rigorous discussions post-annotation.}

As mentioned in \S\ref{sec:stance_categories}, {\stanceosaurus} can also support 3-way stance classification by merging Discussing$_{support}$ and Discussing$_{refute}$ tweets with Supporting and Refuting, respectively. We present the results from BERTweet$_{large}$ for this experiment in Table \ref{tab:3or5-class}. Interestingly, the label F1 for Refuting decreases in the 3-way classification, compared to the 5-way setup. It suggests that identifying the indirect leaning for Discussing$_{refute}$ tweets makes the task harder. Meanwhile, the higher F1 scores for Supporting and Other labels indicate that our classifier is good at detecting tweets that propagate misinformation, even when some of them do not assert a stance explicitly. 

\subsection{Zero-Shot Cross-Lingual Transfer}
\label{subsec:hindi_classification}

Truly multicultural stance identification requires models that are capable of operating across languages. To demonstrate the feasibility of identifying the stance towards misinformation claims in a zero-shot cross-lingual setting, when no training data in the target language is available, we fine-tune models on {\stanceosaurus}' English training set and use all the annotated Hindi/Arabic data as the test set. We experiment with both multilingual BERT \cite{devlin-etal-2019-bert} and XLM-RoBERTa \cite{conneau2020unsupervised}.  Because we  assume no training data is available for the target language, all hyperparameters are tuned on the English dev set. Full results of our 5-class cross-lingual experiments are presented in Table \ref{tab:crosslingual_results}. When trained with class-balanced focal loss, XLM-RoBERTa$_{large}$ achieves 53.1 Macro F1 for Hindi and 50.4 for Arabic, notably outperforming models trained with cross-entropy loss.

\begin{table}[t]
\centering
\small
%\scriptsize
\renewcommand{\arraystretch}{0.85}% Tighter
\resizebox{0.48\textwidth}{!}{

\begin{tabular}{l|ccc}
\toprule
%  \textbf{Model} &  \textbf{Precision} & \textbf{Recall} & \textbf{F1} \\ \midrule
  \multirow{2}{*}{\textbf{Model}} & \multicolumn{3}{c}{\textbf{Stanceosaurus \scalebox{0.75}{(English $\rightarrow$ Hindi)}}} \\
& \hspace*{1mm} \textbf{Precision} & \hspace*{2mm} \textbf{Recall} & \textbf{F1} \\ \midrule
% I used the abbreviations presented in the original XLM-RoBERTa paper
mBERT$_{base}$ + CE &  52.1{\scriptsize $\pm$2.9} & 39.4{\scriptsize $\pm$2.0} & 40.8{\scriptsize $\pm$2.5} \\
+ weighted CE & 55.0{\scriptsize $\pm$ 4.2} & 42.4{\scriptsize $\pm$ 1.4} & 44.3{\scriptsize $\pm$1.8} \\
+ CB$_{\text{foc}}$ & 53.0{\scriptsize $\pm$3.4} & 44.1{\scriptsize $\pm$1.7} & 45.3{\scriptsize $\pm$1.5} \\
\midrule
XLM-R$_{base}$ + CE & 53.2{\scriptsize $\pm$0.1} & 42.6{\scriptsize $\pm$2.1} & 44.3{\scriptsize $\pm$1.9} \\
+ weighted CE & 50.3{\scriptsize $\pm$ 3.2} & 44.4{\scriptsize $\pm$ 1.9} & 44.6{\scriptsize $\pm$1.5} \\
+ CB$_{\text{foc}}$ & 52.8{\scriptsize $\pm$2.1} & 46.5{\scriptsize $\pm$0.7} & 47.4{\scriptsize $\pm$0.9} \\
\midrule
XLM-R$_{large}$ + CE & 55.7{\scriptsize $\pm$3.5} & 49.0{\scriptsize $\pm$1.8} & 49.9{\scriptsize $\pm$1.6} \\
+ weighted CE & \textbf{57.5}{\scriptsize $\pm$ 1.3} & 51.1{\scriptsize $\pm$ 0.9} & 52.5{\scriptsize $\pm$ 1.0} \\
+ CB$_{\text{foc}}$ \vspace*{1mm} & 57.4{\scriptsize $\pm$2.1} & \textbf{51.5}{\scriptsize $\pm$1.3} & \textbf{53.1}{\scriptsize $\pm$1.6}\\
\toprule

%\toprule
%  \textbf{Model} &  \textbf{Precision} & \textbf{Recall} & \textbf{F1} \\ \midrule

  & \multicolumn{3}{c}{\textbf{Stanceosaurus \scalebox{0.75}{(English $\rightarrow$ Arabic)}}} \\
 \midrule
% I used the abbreviations presented in the original XLM-RoBERTa paper
mBERT$_{base}$ + CE &  44.8{\scriptsize $\pm$4.0} & 40.1{\scriptsize $\pm$2.5} & 40.0{\scriptsize $\pm$2.0} \\
+ weighted CE & 44.1{\scriptsize $\pm$ 3.3} & 40.7{\scriptsize $\pm$ 1.6} & 39.7{\scriptsize $\pm$1.7} \\
+ CB$_{\text{foc}}$ & 46.1{\scriptsize $\pm$2.6} & 44.7{\scriptsize $\pm$1.1} & 43.1{\scriptsize $\pm$0.2} \\
\midrule
XLM-R$_{base}$ + CE & 47.6{\scriptsize $\pm$1.8} & 41.9{\scriptsize $\pm$2.1} & 42.6{\scriptsize $\pm$2.2} \\
+ weighted CE & 46.1{\scriptsize $\pm$ 2.0} & 47.9{\scriptsize $\pm$ 2.5} & 46.1{\scriptsize $\pm$2.1} \\
 + CB$_{\text{foc}}$ & 45.8{\scriptsize $\pm$1.7} & 50.0{\scriptsize $\pm$2.2} & 46.4{\scriptsize $\pm$1.6} \\
 \midrule
XLM-R$_{large}$ + CE & 51.4{\scriptsize $\pm$2.7} & 49.2{\scriptsize $\pm$3.4} & {47.7}{\scriptsize $\pm$2.3} \\
+ weighted CE & 49.6{\scriptsize $\pm$ 1.3} & 49.7{\scriptsize $\pm$ 1.7} & 48.2{\scriptsize $\pm$1.4} \\
+ CB$_{\text{foc}}$ & \textbf{51.9}{\scriptsize $\pm$2.0} & \textbf{52.2}{\scriptsize $\pm$2.6} & \textbf{50.4}{\scriptsize $\pm$0.5} \\
\bottomrule
\end{tabular}
}

\caption{Cross-lingual experiments where the models are trained on the English part of {\stanceosaurus} and evaluated on the Hindi/Arabic data. Models trained with class-balanced focal loss (CB$_{\text{foc}}$) outperforms those trained with standard and weighted cross-entropy loss (CE) with higher Macro F1 and lower variance.}
\label{tab:crosslingual_results}
\end{table}

\begin{comment}
We also conduct cross-lingual experiments on Stanceosaurus. Specifically, we evaluate performance on the Hindi data, while training stance classifiers on the English data. Our baseline models include:

\begin{itemize}[leftmargin=18pt]
\itemsep-.5em
\item \textbf{Multilingual BERT} \cite{devlin-etal-2019-bert}: A variation of the BERT model that is traind on a multilingual dataset. 
\item \textbf{XLM-RoBERTa} \cite{conneau2020unsupervised}: Multilingual transformer-based model trained on 100 languages \cite{DBLP:journals/corr/abs-1907-11692} from CommonCrawl. 
% which increases the amount of data for low resource languages. 
\end{itemize}
\end{comment}

%Both models are fine-tuned using cross-entropy and class-balanced focal loss. 
%Even though performance of cross-lingual models on Hindi is lower compared to our best English test set F1, these preliminary results are promising.

%\subsection{Evaluation on RumourEval}
%\label{sec:rumoureval}

\subsection{Combining {\stanceosaurus} + RumourEval}
\label{sec:domain_adaptation_rumoureval}

Because {\stanceosaurus} follows a similar labeling scheme as existing stance corpora, such as RumourEval \cite{gorrell2018rumoureval}, this raises a natural question: is it possible to achieve better performance by combining the two datasets?  

We first confirm that fine-tuning BERTweet$_{large}$ with class-balanced focal loss is also the best performing model on RumourEval-2019's original 4-class evaluation setup, outperforming the weighted cross-entropy loss used in BUT-FIT \cite{fajcik-etal-2019-fit},\footnote{BUT-FIT \cite{fajcik-etal-2019-fit} is one of the state-of-the-art methods on RumourEval-2019, following closely (0.2\% lower Macro F1) behind the winning system BLCU\_NLP \cite{yang-etal-2019-blcu}. BLCU\_NLP achieved a Macro F1 score of 0.62 and used specialized features but is not open-sourced.} as shown in Table \ref{tab:rumoureval_results}. To evaluate cross-dataset performance, we then convert both {\stanceosaurus} and RumourEval-2019 into 3-way stances to minimize the differences between their annotation schemes. RumourEval is converted by collapsing Discussing and Querying instances into the Other category. When merging the datasets, we upsample the RumourEval dataset to twice its size to counteract the imbalance between the two datasets. Table \ref{tab:combined_datasets} shows models trained on in-domain data achieve higher performance than the naive merging of the two datasets for training. %We observed higher variances while training models on RumourEval dataset, similar to \citet{fajcik-etal-2019-fit}. 

\begin{table}[t]
\centering
\small
%\scriptsize
\renewcommand{\arraystretch}{0.85}% Tighter
\resizebox{0.48\textwidth}{!}{
\begin{tabular}{l|ccc}
\toprule
  \multirow{2}{*}{\textbf{Model}} & \multicolumn{3}{c}{\textbf{RumourEval-2019}} \\
&  \textbf{Precision} & \textbf{Recall} & \textbf{F1} \\ \midrule
BERT$_{large}$ + CE &  \textbf{66.8}{\scriptsize $\pm$3.5} & 51.8{\scriptsize $\pm$2.3} & 56.0{\scriptsize $\pm$1.8} \\
+ weighted CE & 61.8{\scriptsize $\pm$4.5} & \textbf{56.7}{\scriptsize $\pm$3.8} & 56.7{\scriptsize $\pm$3.9} \\
+ CB$_{\text{foc}}$ & 62.5{\scriptsize $\pm$6.0} & 54.6{\scriptsize $\pm$1.9} & \textbf{57.5}{\scriptsize $\pm$2.8}\\
 \midrule
BERTweet$_{large}$ + CE & 68.6{\scriptsize $\pm$5.0} & \textbf{62.4}{\scriptsize $\pm$1.3} & 64.0{\scriptsize $\pm$1.5} \\
 + weighted CE & 68.4{\scriptsize $\pm$4.3} &  62.1{\scriptsize $\pm$2.5} &  63.0{\scriptsize $\pm$2.9} \\
 + CB$_{\text{foc}}$ & \textbf{74.4}{\scriptsize $\pm$3.9} & 61.8{\scriptsize $\pm$1.8} & \textbf{65.7}{\scriptsize $\pm$1.4}  \\
\bottomrule
\end{tabular}
}
\caption{Results on RumourEval-2019 that compare different models trained with class-balanced focal loss (CB$_{\text{foc}}$), standard and weighted cross-entropy losses.}
%We show mean and standard deviations of all metrics when trained with 5 random seeds. We achieve new SOTA on RumourEval dataset with BERTweet$_{large}$ model trained when with class-balanced focal loss.
\label{tab:rumoureval_results}
\end{table}

% Finally, they ensemble by training on multiple random seeds. We skip the ensembling part for a fair comparison against our setup.
% What do we want to prove in this experiment? Focal Loss helps more than other weighted Cross Entropy loss used for data imbalance
% Comparison with SOTA BUT_FIT model

%For {\stanceosaurus}, we follow the strategy described in \S \ref{sec:stance_categories}. 

To close this performance gap, we adopt the \textbf{EasyAdapt} \cite{daume2007frustratingly,bai-etal-2021-pre} method to fine-tune BERTweet$_{large}$ on the combination of RumourEval and Stanceosaurus. EasyAdapt creates three identical copies of the contextualized representations of the input, which are concatenated and fed into a linear layer before softmax classification. The parameters in the linear layer that correspond to the first and third copies are updated when training on {\stanceosaurus}, while others are zeroed out; the parameters that correspond to the second and third copies are updated when training on RumourEval. This enables the model to encode representations that are specific to each dataset and domain-independent parameters that can transfer between the two datasets. BERTweet$_{large}$ with EasyAdapt  achieves 67.4 Macro F1 for {\stanceosaurus} and 65.8 Macro F1 for RumourEval, outperforming the in-domain model performance for {\stanceosaurus} and matching the in-domain model performance of RumourEval.

%models trained on in-domain data achieve higher performance.
%We fine-tune BERTweet$_{large}$ using class-balanced focal loss and report performance on {\stanceosaurus} and RumourEval in Table \ref{tab:combined_datasets}. 

% one representing the RumourEval domain, another representing {\stanceosaurus}, and a third domain-independent representation. These contextualized vectors  three times the size of the base model's final classification layer. When encoding data from a specific domain, the other domain's representations are zeroed out, so one of the three vectors is always 0.0.  This enables the model to encode representations that are specific to RumourEval and {\stanceosaurus}, in addition to domain-independent representations that can transfer between the two datasets. 

\begin{table}
\centering
\small
%\scriptsize
%\renewcommand{\arraystretch}{0.85}% Tighter
\resizebox{0.48\textwidth}{!}{
%\resizebox{0.49\textwidth}{!}{
\begin{tabular}{l|ccc|ccc}
\toprule
\multirow{2}{*}{ \textbf{\backslashbox{Train}{Test}}} & \multicolumn{3}{c|}{\textbf{Stanceosaurus}} & \multicolumn{3}{c}{\textbf{RumourEval}} \\
& \textbf{Prec.} & \textbf{Rec.} & \textbf{F1} & \textbf{Prec.} & \textbf{Rec.} & \textbf{F1} \\ \midrule
Stanceosaurus & 66.9 & \textbf{67.1} & 66.8 & 44.8 & 43.8 & 41.2 \\
RumourEval & 39.8 & 43.6 & 40.6 & \textbf{79.6} & 59.7 & 65.7 \\
Combined & 66.6 & 66.0 & 66.2 & 61.1 & 63.4 & 60.6 \\
EasyAdapt & \textbf{68.3} & 67.0 & \textbf{67.4} & 74.4 & \textbf{62.6} & \textbf{65.8} \\
\bottomrule
\end{tabular}
}
\caption{Cross-domain experiments on {\stanceosaurus} and RumourEval. 
We fine-tune BERTweet$_{large}$ using class-balanced focal loss. Performance drops significantly when training on one dataset and testing on the other.
%Although cross-domain performance is signifcantly lower, combining the two datasets with EasyAdapt enhances classification performance in both.
However, with \textbf{EasyAdapt} \cite{daume2007frustratingly,bai-etal-2021-pre}, we attain a single model that achieves best performance on {\stanceosaurus} while being on-par with in-domain RumourEval model performance.
}\label{tab:combined_datasets}
\end{table}

% For all 3 data settings

%Thus, we empirically show that both datasets complement each other in improving overall performance.

% We evaluate the with combining the Stanceosaurus and RumourEval datasets together when training the models. Using the 3-class stance framework, we are able to reasonably combine the two datasets together to get 16132 datapoints. We train BERTweet$_{large}$ with CB focal loss, our best performing model, on just the Stanceosaurus training set, just the RumourEval training set, and the combined training set in the 3-class stance framework. For each training set, we run BERTweet$_{large}$ on both the Stanceosaurus and RumourEval test set.  
% Table \ref{tab:combined_datasets} shows the full results of comparing 3-stance on Stanceosaurus and RumourEval datasets.

\subsection{Stance Detection for Unseen Countries}
The English dataset comprises 97 international and 93 regional claims. We test BERTweet's ability to generalize toward regional claims by training on international claims. Specifically, we create a new train-test-dev split, with 10740/5701/4896 datapoints spread around 97/43/42 claims. Table \ref{tab:unseencountries} shows the results stratified by source. Performance on the regional data varies widely between sources. Poynter and AFP Fact Check New Zealand, two sources with the most international data, have the best F1s at 63.0 and 63.5 respectively. 
\label{sec:unseen_countries}

\begin{table}[t]
\centering
\small
\resizebox{0.48\textwidth}{!}{
\begin{tabular}{l|c|ccc}
\toprule
\textbf{Fact Check Source} & \textbf{\#test} & \hspace*{1mm} \textbf{Precision} & \textbf{Recall} & \hspace*{2mm} \textbf{F1} \\ \midrule
AAP Fact Check & 452 & 50.1 & 39.4 & 40.6 \\
AFP Fact Check CAN & 824 & 71.5 & 54.7 & 58.7 \\
AFP Fact Check NZ & 224 & 64.2 & 63.7 & 63.5 \\
Blackdotresearch & 516 & 65.7 & 62.0 & 60.4 \\
Factly & 447 & 59.4 & 68.2 & 62.5 \\ 
FullFact & 474 & 57.0 & 55.4 & 55.8 \\ 
Poynter & 118 & 73.2 & 61.3 & 63.0 \\ 
Politifact & 614 & 57.7 & 53.7 & 51.8 \\
Snopes & 1402 & 61.4 & 52.0 & 54.4 \\
\midrule
All & 5071 & 62.9 & 54.3 & 57.1 \\
\bottomrule
\end{tabular}
}
\caption{Results on Unseen Countries experiment. BERTweet$_{large}$ finetuned on class-balanced focal loss is trained on international claims and evaluated on regional claims, stratified by fact-checking source. The model achieves an aggregate F1 that is somewhat lower than its counterpart in the Unseen Claims experiment.}
\label{tab:unseencountries}
\end{table}

\section{Conclusion}
% What are the main things we want to cover in conclusion
% What is our dataset and why is it useful
% \wx{need to update} 
We introduce {\stanceosaurus}, a new corpus of 28,033 social media messages annotated with stance towards 251 misinformation claims originating from 15 multicultural fact-checking sources. To the best of our knowledge, {\stanceosaurus} is the largest stance dataset yet. Stanceosaurus contains consistent annotations across claims and languages, and stance classifier models trained on our dataset can perform well on unseen claims and languages.
% Bragging about class-balanced focal loss
Our experiments demonstrate that class-balanced focal loss consistently improves upon cross-entropy loss in addressing the stance label-imbalance issue.
% Experiments with EasyAdapt
Furthermore, the domain adaptation experiments with EasyAdapt show it is possible to utilize RumourEval data to achieve even better performance on {\stanceosaurus} despite significant differences in their data collection strategies.  
% Cross-lingual experiments 
% Cross-lingual experiment using Hindi and Arabic as target languages, showcases the ability of {\stanceosaurus} to facilitate stance prediction without annotations in the target language. 
Our work represents a step towards the development of accurate models that can track the spread of misinformation online across diverse languages and cultures.

\section*{Limitations}
% \wx{need to update} 
We currently use manually curated search queries for collecting tweets related to misinformation claims in {\stanceosaurus}. While we tried our best to include relevant keywords and their synonyms in the search queries, it still requires careful manual effort and may not be exhaustive in finding all relevant tweets related to the claim. Furthermore, it is non-trivial to extend such queries to new claims and languages. Future work could look at automatically generating these queries using a few-shot shot in-context demonstrations with large language models \cite{NEURIPS2020_1457c0d6}. 

We collect the {\stanceosaurus} dataset with all the human resources available to us for three languages. We leave annotations for more languages for future work. We will also release our detailed data annotation guideline and invite other researchers to extend our work to set a standard benchmark for stance classification. 

There are also potential biases in the claims that reflect the biases of content moderators from the fact-checking sources. We made our best effort to identify a list of fact-checking sources based on Wikipedia and pre-existing datasets used in the NLP community to collect claims from different countries and languages. We randomly sample claims from these sources and, since we are constructing a Twitter-based dataset, we are only able to include claims that have been discussed on Twitter. If the claim is unpopular on Twitter, we cannot sufficient data for annotation. Following Twitter's Developer Agreement and Policy, we release our dataset freely for academic research and include the full set of claims in the Appendix of the paper for readers to examine the potential biases in our dataset more conveniently. 

Although the class-balanced focal loss improves stance classification in data imbalanced settings, our models are still far from perfect. We do not use user-specific, temporal, and network features as additional context which has been shown to improve prediction performance \cite{10.1145/3359307, lukasik-etal-2016-hawkes}.

\section*{Broader Impact and Ethical Considerations}

% \wx{update this section; it doesn't will count towards page limit in CFP} 
We will release our dataset under Twitter Developer Agreement,\footnote{\url{https://developer.twitter.com/en/developer-terms/agreement-and-policy}} which grants permissions for academic researchers to share Tweet IDs and User IDs for non-commercial research purposes, as of October 1st. 2022.

Our datasets and models are developed for research purposes and may contain unknown biases towards certain demographic groups or individuals \citep{sap2019risk}.  Further investigation into systematic biases should be conducted before deployment in a production environment.

Social media companies currently struggle with content moderation in non-Western countries.\footnote{\url{https://www.washingtonpost.com/technology/2021/10/24/india-facebook-misinformation-hate-speech/}} We hope \stanceosaurus{} will help stimulate more public research that can help shed light on how to inhibit the spread of dangerous misinformation across languages and cultures.

\section*{Acknowledgments}

% The acknowledgments should go immediately before the references. Do not number the acknowledgments section.
% \textbf{Do not include this section when submitting your paper for review.}

We thank three anonymous reviewers for their helpful comments. We also thank Chao Jiang for providing his codebase; Chase Perry, Mohamed Ghanem, Angana Borah, Rucha Sathe, Andrew Duffy, and Kenneth Koepcke for their help with data annotation. This research is supported in part by NSF awards IIS-2144493 and IIS-2052498, in addition to ODNI and IARPA via the BETTER program (contract 19051600004). The views and conclusions contained herein are those of the authors and should not be interpreted as necessarily representing the official policies, either expressed or implied, of NSF, ODNI, IARPA, or the U.S. Government. The U.S. Government is authorized to reproduce and distribute reprints for governmental purposes notwithstanding any copyright annotation therein.
\bibliography{anthology,custom}
\bibliographystyle{acl_natbib}

\appendix

\section{Customized Queries for Retrieving Tweets}
\label{appendix:claims_and_queries}
We present example claims and their search queries from each of the three languages in Table \ref{tab:image_claim_examples}.

\begin{table*}[t]
\centering
% \small
\scriptsize
\renewcommand{\arraystretch}{0.85}% Tighter
\resizebox{\textwidth}{!}{
\begin{tabular}{ll|ccc|ccc||ccc}
\toprule
\hspace*{10mm} & \multirow{2}{*}{  \textbf{\backslashbox{Train}{Test}}} & \multicolumn{3}{c|}{\textbf{Root Tweets (56.1\%)}} & \multicolumn{3}{c||}{\textbf{Reply Tweets (43.9\%)}} & \multicolumn{3}{c}{\textbf{All Tweets}}\\
& & \textbf{Precision} & \textbf{Recall} & \textbf{F1} & \textbf{Precision} & \textbf{Recall} & \textbf{F1} & \textbf{Precision} & \textbf{Recall} & \textbf{F1} \\ \midrule
%\multicolumn{2}{l|}{Stanceosaurus} &  &  &  &  &  & &  &  & \\
\multicolumn{2}{l|}{$\smalltriangleright$ root} & \textbf{60.8}{\tiny $\pm$1.2} & 63.0{\tiny $\pm$1.8} & \textbf{61.7}{\tiny $\pm$1.3} & 35.1{\tiny $\pm$1.2} & 40.4{\tiny $\pm$2.1} & 35.4{\tiny $\pm$1.5}  & 52.4{\tiny $\pm$0.4} & 56.1{\tiny $\pm$1.4} & 53.6{\tiny $\pm$0.9} \\
\multicolumn{2}{l|}{$\smalltriangleright$ reply w/o context} & 53.7{\tiny $\pm$3.2} & 42.1{\tiny $\pm$4.2} & 43.9{\tiny $\pm$4.7} & 33.1{\tiny $\pm$1.5} & 35.3{\tiny $\pm$2.0} & 33.2{\tiny $\pm$1.3} & 44.7{\tiny $\pm$2.9} & 39.7{\tiny $\pm$3.4} & 40.5{\tiny $\pm$3.3} \\
\multicolumn{2}{l|}{$\smalltriangleright$ reply w/ context} & 49.0{\tiny $\pm$3.2} & 37.3{\tiny $\pm$0.7} & 38.6{\tiny $\pm$1.5} & 42.6{\tiny $\pm$3.3} & 41.3{\tiny $\pm$2.4} & 41.6{\tiny $\pm$2.5} & 47.7{\tiny $\pm$3.7} & 39.1{\tiny $\pm$1.4} & 41.0{\tiny $\pm$2.2} \\
\multicolumn{2}{l|}{$\smalltriangleright$ root + reply w/o context} & 60.2{\tiny $\pm$1.4} & 63.2{\tiny $\pm$1.7} & 61.4{\tiny $\pm$0.8} & 35.7{\tiny $\pm$0.9} & 42.1{\tiny $\pm$1.1} & 36.9{\tiny $\pm$0.9} & 52.1{\tiny $\pm$1.1} & 56.6{\tiny $\pm$1.2} & 53.8{\tiny $\pm$0.7} \\
\multicolumn{2}{l|}{$\smalltriangleright$ root + reply w/ context} & 60.3{\tiny $\pm$ 2.0} & \textbf{63.5}{\tiny $\pm$ 1.8} & 61.5{\tiny $\pm$ 1.5} & \textbf{56.7}{\tiny $\pm$ 4.0} & \textbf{46.2}{\tiny $\pm$ 1.8} & \textbf{49.5}{\tiny $\pm$ 2.0} & \textbf{66.0}{\tiny $\pm$0.4} & \textbf{65.5}{\tiny $\pm$1.1} & \textbf{65.6}{\tiny $\pm$0.7} \\
\bottomrule
\end{tabular}
}
\caption{Ablation experiments to study the impact of context in 5-way stance classification. In particular, we split the Twitter threads within {\stanceosaurus}' training set into root tweets (those with no parent in the conversation thread) and reply tweets (tweets that are written in response to another message). In all experiments, we train BERTweet$_{large}$ using cross-entropy loss. Results suggest that predicting the stance of reply tweets is significantly harder than root tweets. Context improves the overall stance classification performance mainly by improving prediction on reply tweets.
}\label{tab:context_results} 
\end{table*}

\begin{table}[t]
\centering
\small
\resizebox{0.48\textwidth}{!}{
\begin{tabular}{l|cc|c||c}
\toprule
\multirow{2}{*}{\textbf{Fact Check Source}} & \multicolumn{3}{c||}{\textbf{Unseen Source}} & \multicolumn{1}{c}{\textbf{Unseen Claims}}
\\
% & \textbf{Train$_{removed}$} & \textbf{F1$_{removed}$} & \textbf{F1} & \\ \midrule
 & \textbf{\#train$^*$} & \textbf{\#test}& \multicolumn{1}{c||}{\textbf{F1}} & \textbf{F1} \\ \midrule

AAP Fact Check & 11135 & 477  & 58.7 & \textbf{59.2} \\
AFP Fact Check CAN & 10941 & 459 & 57.6 & \textbf{59.6} \\
AFP Fact Check NZ & 10554 & 482 & 54.5 & \textbf{55.7} \\
Blackdotresearch & 10699 & 517 & 57.3 & \textbf{59.3} \\
Factly & 10693 & 318 & 59.4 & \textbf{61.4} \\ 
FullFact & 10783 & 602 & 60.0 & \textbf{61.8} \\ 
Politifact & 10927 & 838 & 50.6 & \textbf{60.0} \\
Poynter & 10715 & 321 & 52.4 & \textbf{55.3} \\ 
Snopes & 10593 & 736 & 57.7 & \textbf{58.0} \\
\midrule
All & 12130 & 4750 & - & \textbf{61.0} \\
\bottomrule
\end{tabular}
}
\caption{
Results of BERTweet$_{large}$ with the class-balanced focal loss on unseen fact-checking sources. For each source, we remove associated tweets from train/dev in {\stanceosaurus}' standard data split. Macro F1 scores are computed on a subset of the test set with tweets only from the unseen source. We also report the performance of the 
same model trained on full train/dev splits in {\stanceosaurus} with tweets from all sources. Performance is degraded when predicting stance on unseen sources, but not by a large margin.
% Results of BERTweet$_{large}$ fine-tuned using class-balanced focal loss on unseen fact-checking sources. For each source, we remove all tweets from the train and dev splits. Macro F1 score is computed on a subset of the test set that consists of tweets from the source that is unseen during training. We also report the performance of the best model trained on the {\stanceosaurus}' standard data split, which includes tweets from the same source. Performance is degraded when classifying stance on unseen sources, but not by a large margin.
%Results of BERTweet$_{large}$ stance classifier trained with the class-balanced focal loss on unseen sources. We convert every source into an unseen source by removing its tweets from train and dev data and creating a new data subset (\textbf{Train$^*$}). We evaluate the Macro F1 performance on the test set tweets exclusively from the unseen source (\textbf{Test$^*$}) and also compare against the performance of the best model trained on the unseen claims data split of {\stanceosaurus}. In general, the absence of unseen source in the training data hurts performance on unseen test tweets, although not by a huge margin.
}
\label{tab:unseensources}
\end{table}

\section{Stance Classification with Context}

\subsection{Annotation Example of Tweets in Reply Chain}
\label{app:example_replychain}

Table \ref{tab:tweet_examples} shows representative examples of different stances towards the claim {\em ``The COVID-19 Vaccine will make your body magnetic''}. Note that some tweets are context-dependent (e.g., {\em ``No that is not true''}); their stance can only be determined with appropriate context.

\subsection{Guidelines for Tricky Annotation}
\label{subsec:tricky_anno}

We identified some common scenarios in our annotation which lead to annotation disagreements in our preliminary analysis of the data. We designed specific guidelines to improve annotation consistency, including:
\begin{itemize}
    \item If the claim has a lot of information, we should focus on the core contentious part of the claim when judging the stance of the tweets.
    \item If the tweet is giving an analysis of the contentious event or talking about an adjacent event (regional) then it should be considered Discussing.
    \item If the tweet is just emojis, praise, or pleasant message (e.g., ``\textit{thank you}'', ``\textit{good job sir}'') towards a context tweet, consider it Discussing with the leaning inherited from the Stance of the context tweet.
    \item For querying, the tweet should be questioning the veracity of the claim and not any other question about the incident.
    \item If the main purpose of the tweet is gauging the people’s opinions related to the claim then it is Discussing.
    \item If the tweet is posing a question with \#fakenews or \#factcheck but the URL asserts that the claim is fake then it should be judged Refuting. However, if the URL is also a question without a judgment then it should be considered Discussing.
    \item If a reply tweet is adding information/opinion on top of the context (assuming that the context tweet is true) then annotate Discussing with Leaning inherited from the context.
\end{itemize}

% \wx{This table primarily show the impact of including context (presumably it helps); explain what is ``context'' in the table caption; need to carefully decide whether we should do 3-class or 5-class, and which model we should use for this table.}

\begin{table*}[t]
\scriptsize
\centering

\resizebox{\textwidth}{!}{%
\begin{tabular}{p{9.5cm}p{6cm}}
\toprule
\multicolumn{1}{c}{\textbf{Claim}} & \multicolumn{1}{c}{\textbf{Query}} \\ \midrule
Easter is a celebration for the Mediterranean Goddess Ishtar & (easter ishtar) lang:en -filter:retweets  \\ \midrule
The false positive rate for a COVID-19 test is very high & ((COVID OR coronavirus) AND false positive) lang:en -filter:retweets  \\ \midrule
% \multicolumn{2}{c}{\includegraphics[width=1.0\textwidth]{images/hindi_claim1.pdf}} \\
% \textit{Translation:} Sharing some pictures and videos, it is being claimed that a Christian priest appointed in a temple in Tamil Nadu insulted the idols by drinking alcohol in the temple. & \textit{Translation:} christian priest temple alcohol lang:hi since:2021-11-10  -filter:retweets \\ \midrule
% \multicolumn{2}{c}{\includegraphics[width=1.0\textwidth]{images/hindi_claim2.pdf}} \\
% \textit{Translation:} Degrees of 100 girl students of Srinagar Medical College were canceled because they raised the slogan 'Pakistan Zindabad'. & \textit{Translation:} Srinagar degree cancel pakistan lang:hi since:2021-10-22 -filter:retweets\\
\multicolumn{2}{c}{\includegraphics[width=1.0\textwidth]{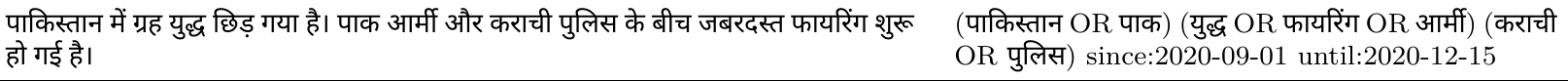}} \\
\textit{Translation:} In-house war has broken out in Pakistan. Heavy firing has started between Pak Army and Karachi Police. & \textit{Translation:} (pakistan OR pak) (war OR firing OR army) (karachi OR police) since:2020-09-01 until:2020-12-15\\
\midrule
\multicolumn{2}{c}{\includegraphics[width=1.0\textwidth]{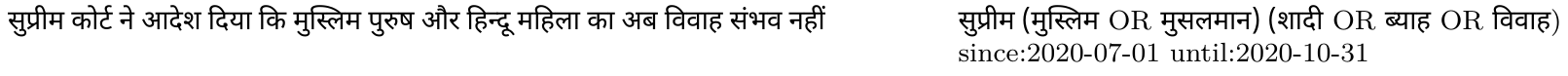}} \\
\textit{Translation:} Supreme Court orders that marriage of Muslim man and Hindu woman is no longer possible. & \textit{Translation:} Supreme (muslim OR musalmaan) (marriage OR wedding OR matrimony) since:2020-07-01 until:2020-10-31\\
\midrule

\multicolumn{2}{c}{\includegraphics[width=1.0\textwidth]{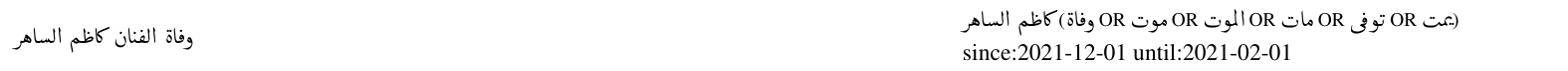}} \\

\textit{Translation:} Death of the artist Kadim Al Sahir & \textit{Translation:} (Kadim AND Al Sahir) (Death OR Die OR Died)  since:2021-12-01 until:2022-02-01\\

\midrule

\multicolumn{2}{c}{\includegraphics[width=1.0\textwidth]{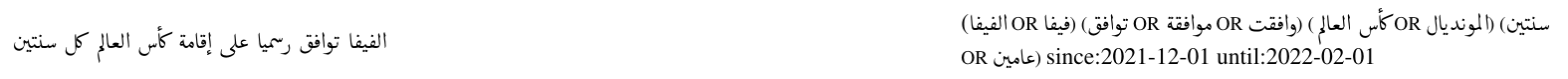}} \\

\textit{Translation:} FIFA officially agrees to host the world cup every two years & \textit{Translation:} FIFA (agrees OR agreed) (world cup) (two years) since:2021-12-01 until :2022-02-01\\

\midrule

\end{tabular}
}
\caption{Example English and Hindi claims with corresponding search queries. Queries are manually constructed to cast a broad net, retrieving both relevant and irrelevant messages containing the keywords.}
\label{tab:image_claim_examples}
\end{table*}

\begin{table*}[t]
\small
\centering
% \resizebox{0.99\textwidth}{!}{
% \begin{tabular}{p{0.15\textwidth}p{0.7\textwidth}p{0.15\textwidth}}
% \toprule
% Error Type & Tweet & Pred.\\\midrule
% Inference Error & WH says Trump spoke with Boris Johnson and "wished him a speedy recovery" after the British PM tested positive for coronavirus. & the British PM ()\\\midrule
% Segmentation Error & Washing your hands properly can prevent the spread of diseases such as the \#flu or \#COVID19. See below for the proper way to wash your hands. & your hands\\\midrule
% Ambiguous Case &  & \\
% \bottomrule
% \end{tabular}
% }
\renewcommand{\arraystretch}{0.7}% Tighter
\begin{tabular}{P{0.96\textwidth}}
\toprule
\multicolumn{1}{c}{\fontsize{8}{9.5}\selectfont \textbf{Claim: \textit{The COVID-19 Vaccine has magnets or will make your body magnetic} }}\\\midrule
{$\star$ \fontsize{7.5}{8}\selectfont  \sethlcolor{lightgrey}\textbf{\textcolor{darkgrey}{\hl{Irrelevant}}}: {\it @dbongino is right. you can't tell people to wear a mask if the vaccines work. its like trying to put a north end of a magnet and trying to connect it to a north end another magnet., it will never work. \#foxandfriends
}} \\%\cmidrule{1-1}
\midrule
{$\star$ \fontsize{7.5}{8}\selectfont \sethlcolor{pink}\textbf{\textcolor{darkred}{\hl{Supporting}}}: {\it a friends family member got the covid vaccine and now she can put a magnet up to the injection site and the magnet stays on her arm. }} \\\cmidrule{1-1}

\begin{tabular}{P{0.1cm}P{0.88\textwidth}}
{\hspace{1pt}} & $\drsh$
{\fontsize{7.5}{8}\selectfont \sethlcolor{pink}\textbf{\textcolor{darkred}{\hl{Supporting (only in context)}}}: {\it @ThisIsTexasFF Nano probes / tech / dust.}}
\end{tabular} \\%\cmidrule{1-1}
\midrule
{\fontsize{7.5}{8}\selectfont \sethlcolor{lightblue}\textbf{\textcolor{blue}{\hl{Refuting (only in context)}}}: {\it @Newsweek Why the hell would they even bother with a high quantity of metal in the injection? And the amount that would be required to hold a magnet in place would be ridiculous.}} \\\cmidrule{1-1}

\begin{tabular}{P{0.1cm}P{0.90\textwidth}}
{\hspace{1pt}} & $\drsh$
{\fontsize{7.5}{8}\selectfont \sethlcolor{lightblue}\textbf{\textcolor{blue}{\hl{Refuting (only in context)}}}: {\it
@pentatonicScowl @Newsweek I imagine the people making the claims don't fully understand how magnets work}} %\cmidrule{1-1}
\end{tabular} \\\cmidrule{1-1}

\begin{tabular}{P{0.8cm}P{0.76\textwidth}}
{\hspace{1pt}} & $\drsh$
{\fontsize{7.5}{8}\selectfont \sethlcolor{pink}\textbf{\textcolor{darkred}{\hl{Supporting (only in context)}}}: {\it @AuracleDMG @pentatonicScowl @Newsweek Laugh now, cry later..}}
\end{tabular}\\\cmidrule{1-1}

\begin{tabular}{P{1.5cm}P{0.82\textwidth}}
{\hspace{1pt}} & $\drsh$ $\star$
{\fontsize{7.5}{8}\selectfont \sethlcolor{lightblue}\textbf{\textcolor{blue}{\hl{Refuting}}}: {\it @cis\_kale Your point being? Even if these RNA vaccines contained ferric nanoparticles, they would not be in high enough concentrations to be able to hold a magnet in place. I suspect that blood itself has a higher concentration of ferric particles than the vaccine described in this paper}}
\end{tabular}\\%\cmidrule{1-1}

%\midrule
%{$\star$ \fontsize{7.5}{8}\selectfont {\it  \sethlcolor{lightgreen}\textbf{\textcolor{darkgreen}{\hl{Discussing:}}}@paulajaynebyrne There's quite an overlap between Covid deniers and those who think the vaccine is a conspiracy to change our DNA, implant us with microchips etc. so I think they will refuse it.}}\\%\cmidrule{1-1}
\midrule
{$\star$ \fontsize{7.5}{8}\selectfont \sethlcolor{lightpurple}\textbf{\textcolor{darkpurple}{\hl{Querying}}}: {\it There is a \#covid19 vaccine magnet test circulating on Tiktok, Is it really a thing?!!}} \\\cmidrule{1-1}
\begin{tabular}{P{0.1cm}P{0.88\textwidth}}
{\hspace{1pt}} &
{$\drsh$ \fontsize{7.5}{8}\selectfont \sethlcolor{pink}\textbf{\textcolor{darkred}{\hl{Supporting (only in context)}}}: {\it @Thepurplelilac well, 4 friends out of 9 can stick magnets to their arms so yeah, it's a thing}}
\end{tabular}\\%\cmidrule{1-1}
\midrule
{$\star$ \fontsize{7.5}{8}\selectfont  \sethlcolor{lightgreen}\textbf{\textcolor{darkgreen}{\hl{Discussing}}}: {\it @heggzigu @htmdnl too early to make any presumptions on either side. the truth has a way of exposing itself given enough time. bring a magnet to your vaccination appointment, see how the vaccine reacts with the magnet, maybe even bring a metal detector as well. would that convince you?}}\\\midrule
{$\star$ \fontsize{7.5}{8}\selectfont  \sethlcolor{lightgreen}\textbf{\textcolor{darkgreen}{\hl{Discussing}}}: {\it Fauci: No Concern About Number of People Testing Positive After COVID-19 Vaccine. Spike Protein Vax is magnet for coronavirus. Originally used as turbo booster mounted on virus but too flimsy. Now injected in target in advance of infection, death rate 4X.}}
\\\bottomrule
\end{tabular}
\caption{An example claim and its corresponding tweets from the 5 stance categories (best view in color):
\sethlcolor{lightgrey}\textcolor{darkgrey}{\textbf{\hl{Irrelevant}}}, \sethlcolor{lightblue}\textbf{\textcolor{blue}{\hl{Refuting}}}, \sethlcolor{pink}\textbf{\textcolor{darkred}{\hl{Supporting}}}, 
\sethlcolor{lightgreen}\textcolor{darkgreen}{\textbf{\hl{Discussing}}},
and \sethlcolor{lightpurple}\textbf{\textcolor{darkpurple}{\hl{Querying}}}. $\star$ symbol indicates the tweets we directly retrieved from our query keyword search method. Indented lines with $\drsh$ are replies to parent tweets. }
% HACK to push the table to the top of the page even with no text present
% \vspace{128in}
\label{tab:tweet_examples}
\end{table*}

\subsection{Importance of Considering Context}
\label{subsec:context_ablation}
Stance that is realized in social media messages often depends on the context of a conversation, or links to external webpages, as discussed in \S \ref{sec:retrieving}. In this section, we evaluate the impact of context in the form of parent tweets and URL titles. To ablate context, we first organize tweets in the training data into reply chains. Next, we separate threads into \textit{root tweets} that have no parent in the conversation thread and \textit{reply tweets} that are written in response to another message. 
We fine-tune BERTweet$_{large}$ on (1) only root tweets, (2) only reply tweets, and (3) both root and reply tweets. We also measure the impact of training with and without context. We use standard cross-entropy loss for this comparison study, excluding the impact of hyperparameter choices in the focal loss, as the stance distribution differs between root and reply tweets.

The results in Table \ref{tab:context_results} demonstrate that root tweets, reply tweets, and context are complementary for achieving the best overall performance. The F1 score on root tweets is significantly higher than on reply tweets, indicating the difficulty to determine stance in extended conversations. Unsurprisingly, training only on root tweets achieves a higher 61.7 F1 on root tweets but a lower 35.4 F1 on reply tweets. For models trained only on reply tweets, including context improves performance on reply tweets but hurts performance on root tweets.  %We hypothesize root tweets without context are out-of-domain for the model trained exclusively on reply tweets with context. 

\begin{comment}
We look at the impact of including context and segmenting the dataset based on root and reply tweets. We experiment with testing on root and reply tweets by training models on just the root tweets, just the reply tweets, just the reply tweets without the context appended, the entire dataset without the appended context, and the entire dataset. \newline
For all the models, the performance on root tweets significantly outperforms the performance on reply tweets, indicating the difficulty in classifying long extended Tweet chains. BERTweet$_{large}$ is able to achieve 62.7 F1, the best performance, on the root tweets when trained on root tweet. On the other hand, training on just the reply tweets leads to significant decreases in model performance. BERTweet$_{large}$ performs the best on reply tweets and on all tweets when trained on the entire training set with the appended context. Interestingly, training on the entire dataset without context leads to a lower performance compared to just training on the root tweets. Thus, omitting context from context examples actively harms performance, and the inclusion of context as additional information leads to an increase in performance of the models. Table \ref{tab:context_results} shows the full results of the context ablation studies.
\end{comment}

\subsection{Unseen Fact-checking Sources}
\label{sec:unseen_sources}

Since the claims in {\stanceosaurus} are collected from multicultural sources, we also test stance classifier's performance towards claims found in fact-checking sources that are unseen in the training data.
% In this section, we evaluate the stance classifier's generalization to unseen sources in {\stanceosaurus}.
Specifically, we convert each fact-checking website in {\stanceosaurus} into an unseen source by creating a new data-split and removing its tweets from the train and dev sets.
% For each fact-checking source in {\stanceosaurus}, we create a new data split by excluding its tweets from train and dev set. 
Then, a model trained on this restricted data is evaluated on the test tweets from the selected unseen source. For comparison, we also report the performance of the best model from the unseen claims experiment (\S \ref{sec:unseen_claims} where claims from each source are split into train/dev/test) on these test tweets from the unseen source. 
For every unseen source, we train a BERTweet$_{large}$ stance classifier with class-balanced focal loss and report its results in Table \ref{tab:unseensources}. The models perform worse when the source is removed from training data, with Politifact showing the biggest drop in performance from 60.0 F1 to 50.6 F1. This highlights the importance of source-specific data in classifying misinformation claims.

\section{Additional Details on Conversation Threads}
\label{app:conversation_retrieval}

For each claim from English and Hindi sources, we randomly sample up to 150 tweets for annotation: max 50 tweets (average 50 for English and 48.1 for Hindi) retrieved from our queries, max 50 parent tweets (average 30.7 for English and 8.6 for Hindi), and max 50 children tweets (average 28.3 for English and 33.0 for Hindi) from reply chains. For Arabic, we annotated all the tweets (average 175.8 per claim) retrieved from the search and reply chain. Finally, we organize every tweet such that its immediate parent serves as the context. For tweets containing URLs, we also additionally include the HTML `Title' tag extracted from the URL. About 40.5\% of all tweets in our dataset have a parent tweet in context, while 19.5\% of tweets have associated HTML titles. 

\section{\textit{Stanceosaurus} Claims}
\label{sec:claims}
\begin{figure*}[h]
    \centering
    \includegraphics{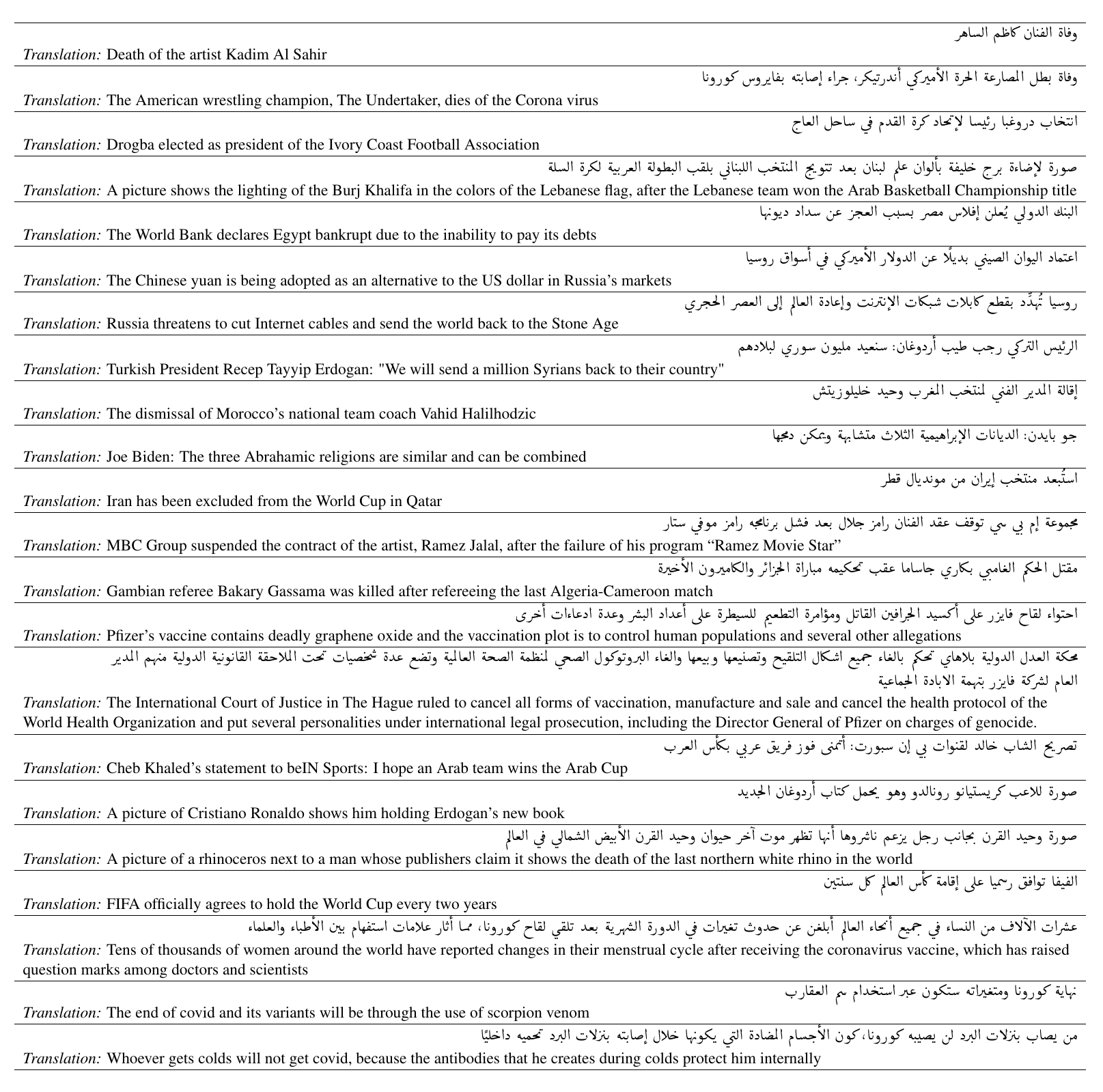}
    \caption{Arabic Claims}
    \label{fig:arabic_label}
\end{figure*}

We provide the full set of English, Hindi, and Arabic claims, with English translations for Hindi and Arabic. Figure \ref{fig:arabic_label} provides all 22 Arabic claims, Table \hyperlink{hindi_claims}{15} provides all 39 Hindi claims, and Table \hyperlink{english_claims}{16} provides all 190 English claims.

\begin{figure*}[h]
    \centering
    \hypertarget{hindi_claims}{}
    \includegraphics{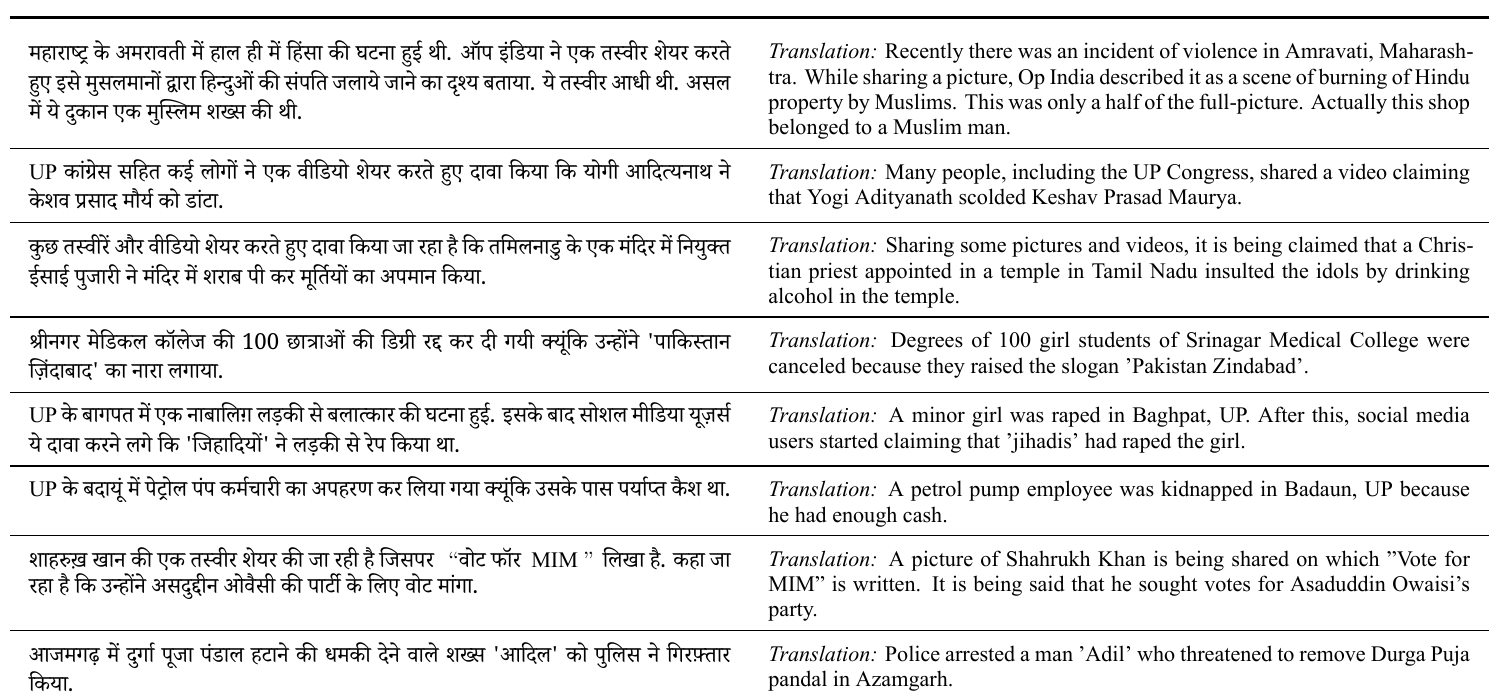}
\end{figure*}

\begin{table*}[h]
    \centering
    \includegraphics{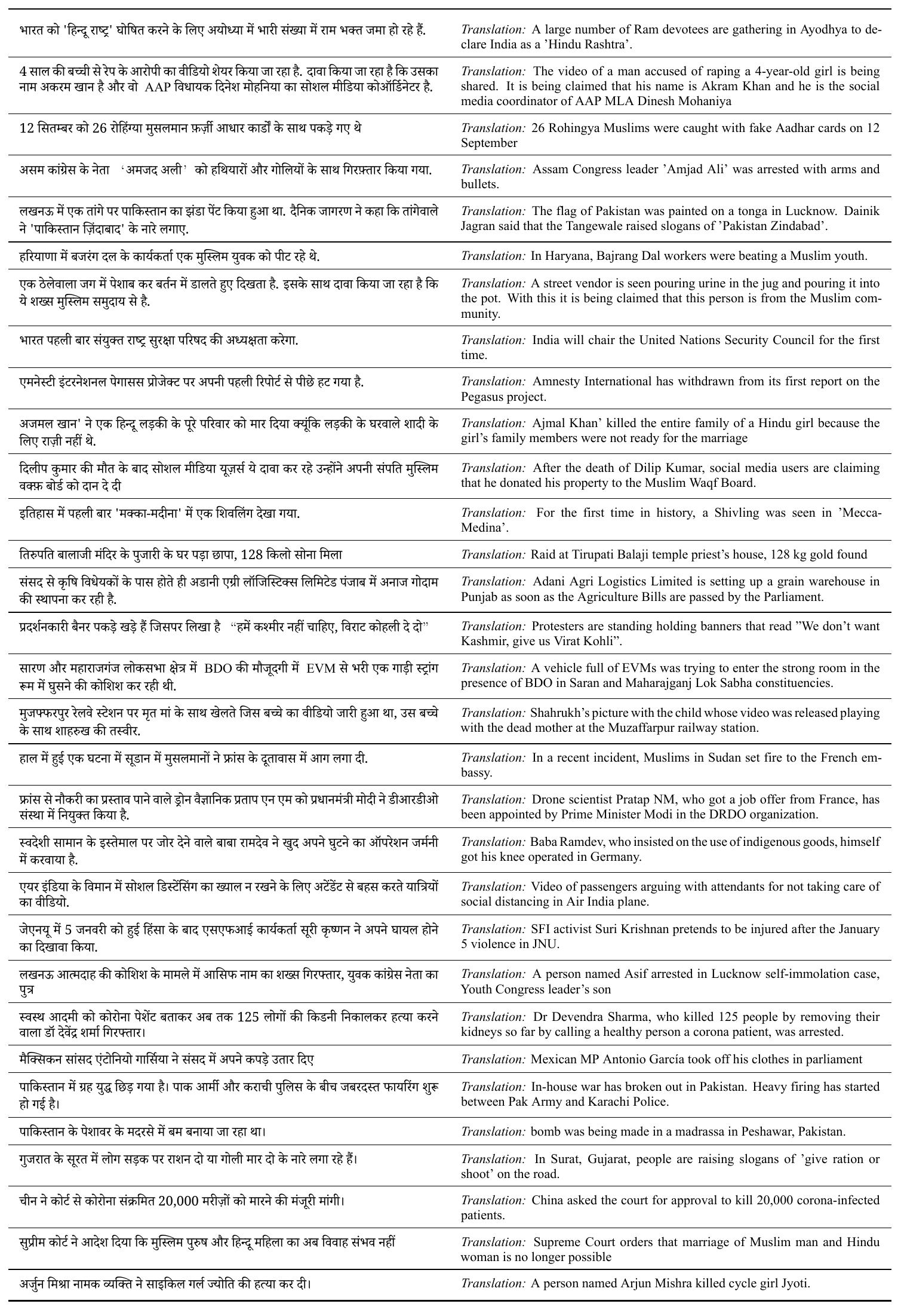}
    \caption{Hindi Claims}
    \label{tab:my_hindi_label_one}
\end{table*}

\begin{figure*}[h]
    \includegraphics{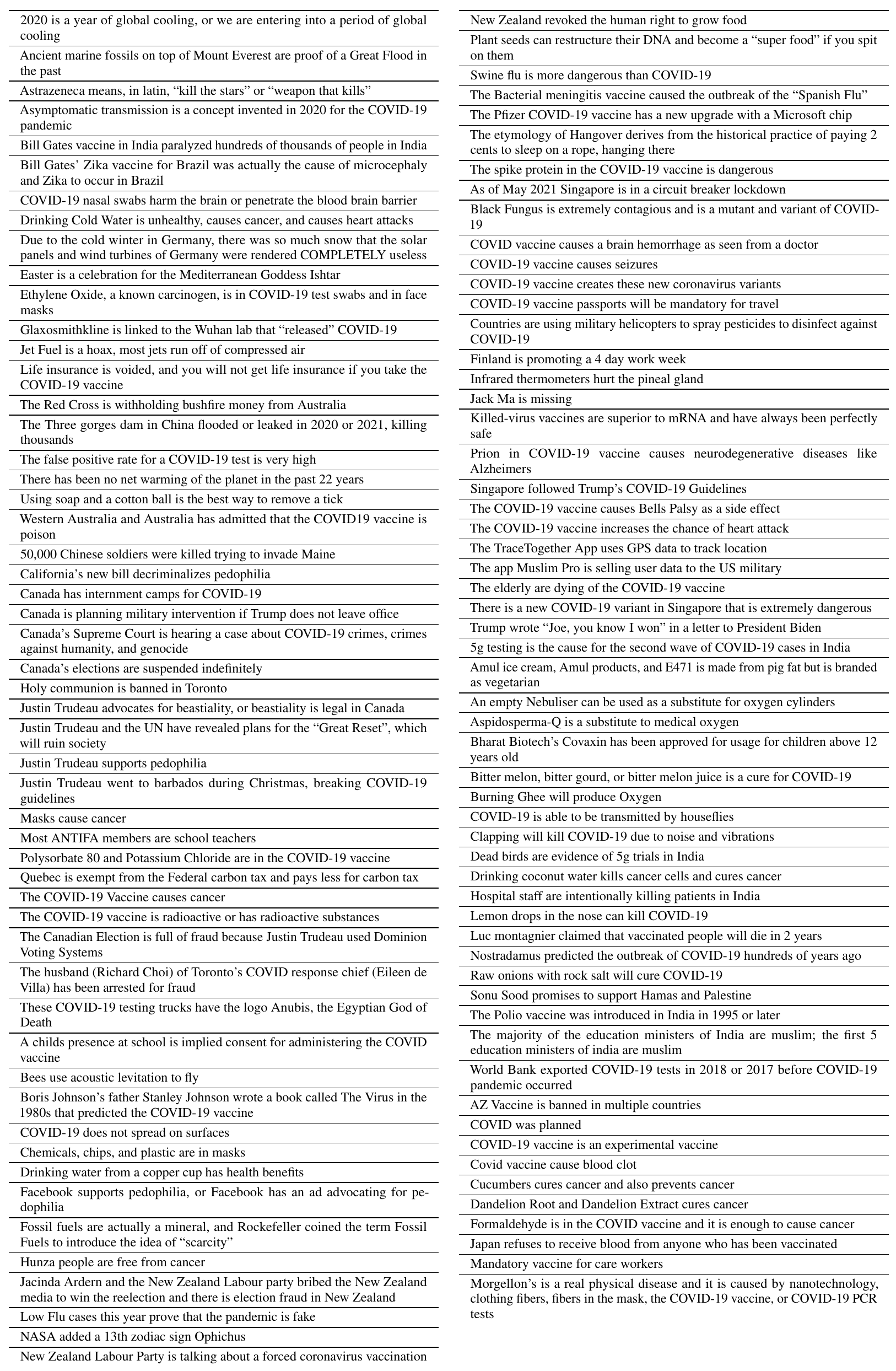}
    \label{fig:my_english_label}
\end{figure*}

\begin{figure*}[h]
    \hypertarget{english_claims}{}
    \includegraphics{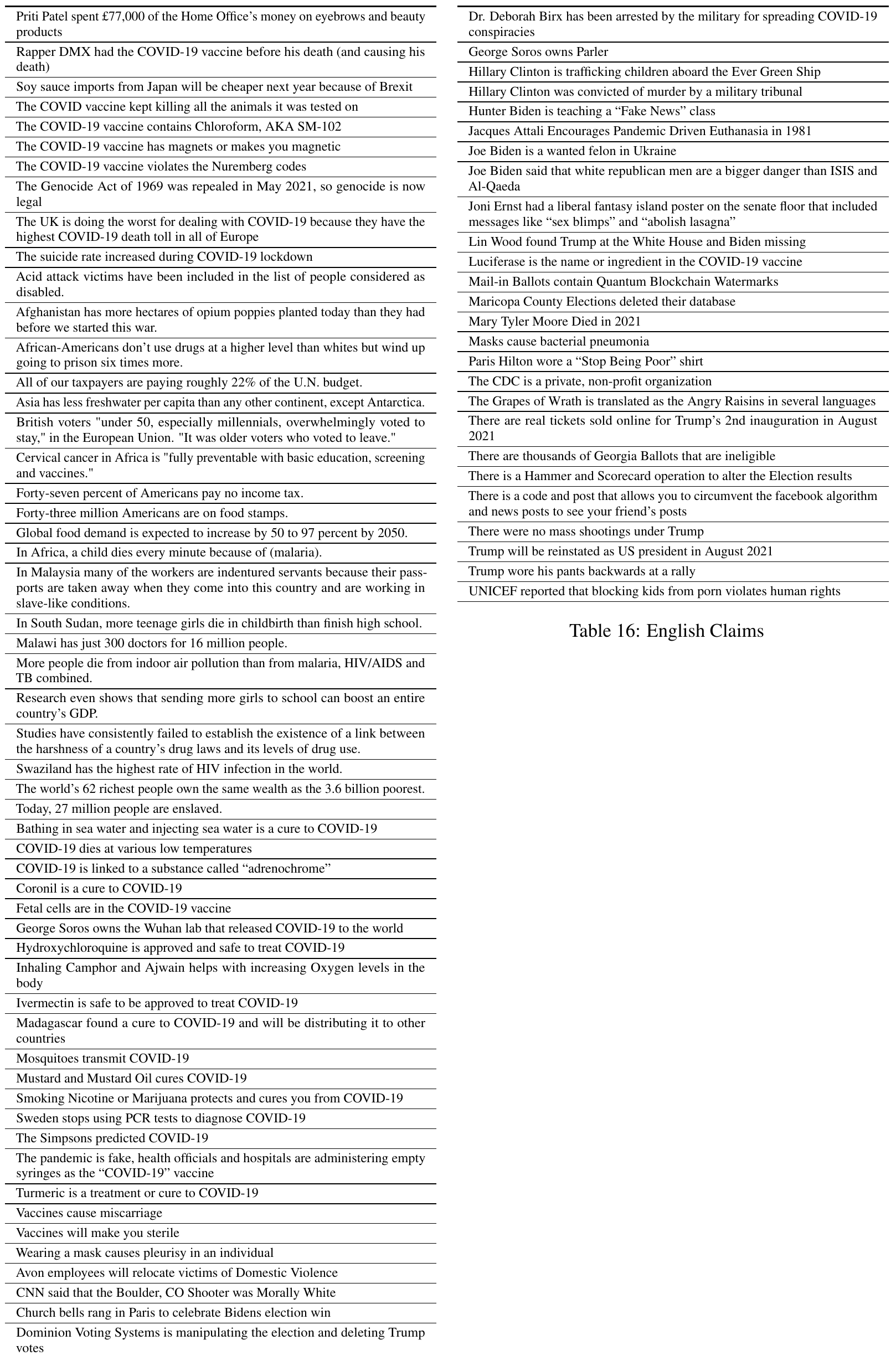}
\end{figure*}

\end{document}